%% file: main.tex
\documentclass[12pt,a4paper]{article} 
\usepackage[left=2.5cm,top=3cm,right=2.5cm,bottom=3cm,bindingoffset=0.5cm]{geometry}

\usepackage{fontspec}
\usepackage{xeCJK}

\usepackage{caption}
\newcommand{\tablesize}{\fontsize{11}{12}} 
\usepackage{subcaption}

\usepackage{fancyhdr}
\usepackage{endnotes} 
\let\footnote=\endnote
\renewcommand{\notesname}

\usepackage{pdflscape}
\usepackage{newtxtext,newtxmath} 
\usepackage{bm} 
\usepackage{booktabs} 
\usepackage{tipa} 
\usepackage{graphicx}
\usepackage{float}
\usepackage{adjustbox}
\usepackage{rotating} 
\usepackage{xcolor}
\usepackage[]{hyperref}
\usepackage{comment}

\usepackage[backend=biber,natbib=true, style=apa]{biblatex}
\usepackage{enumitem}
\usepackage{setspace}
\usepackage{hanging} 
\DeclareDelimFormat{finalnamedelim}{\addspace and\space}
\DeclareDelimFormat{multinamedelim}{\addcomma\space}

\newcommand{\mytitle}[1]{%
  \vspace*{3\baselineskip} 
  \begin{center}
    {\fontsize{12}{14}\selectfont \MakeUppercase{#1}} 
  \end{center}
}

\newcommand{\myauthor}[3]{%
  \begin{center}
    {\fontsize{10.2}{12}\selectfont \textbf{#1}}\\[0.5\baselineskip] 
    {\fontsize{10.2}{12}\selectfont #2}
    \ifx&#3&{}
    \else
      \footnote{#3}
    \fi
  \end{center}
}

\newcommand{\englishabstract}[1]{%
  \vspace{2\baselineskip} 
  {\noindent ABSTRACT} 
  \par
  \vspace{0.5\baselineskip} 
  {\fontsize{10.2}{14}\selectfont #1} 
  \par
}

\newcommand{\englishkeywords}[1]{%
  \vspace{\baselineskip} 
  {\noindent KEYWORDS}  
  \par
  \vspace{0.3\baselineskip} 
  {\fontsize{10.2}{14}\selectfont 
  \def\kw##1{\textbf{##1}}
  #1
  }
  \par
}

\newcommand{\myctitle}[1]{%
  \vspace*{3\baselineskip} 
  \begin{center}
    \fontsize{12}{14}\selectfont \textbf{#1} 
  \end{center}
}

\newcommand{\mycauthor}[3]{%
  \begin{center}
    \fontsize{10.2}{12}\selectfont \textbf{#1}\\[0.5\baselineskip] 
    \fontsize{10.2}{12}\selectfont #2 
    \ifx&#3&{}%
    \else
      \footnote{#3}
    \fi
  \end{center}
}

\newcommand{\chineseabstract}[1]{%
  \vspace{2\baselineskip} 
  {\noindent \textbf{摘要}} 
  \par
  \vspace{0.5\baselineskip} 
  {\fontsize{10.2}{14}\selectfont #1} 
  \par
}

\newcommand{\chinesekeywords}[1]{%
  \vspace{\baselineskip} 
  {\noindent \textbf{关键词}} 
  \par
  \vspace{0.3\baselineskip} 
  {\fontsize{10.2}{14}\selectfont #1} 
  \par
}

\begin{document}



\mytitle{Form and meaning co-determine the realization of tone in Taiwan Mandarin spontaneous speech: the case of T2-T3 and T3-T3 tone sandhi}

\myauthor{Yuxin Lu}{\textit{University of Tübingen}}{}
\myauthor{Yu-Ying Chuang}{\textit{National Taiwan Normal University}}{}
\myauthor{R. Harald Baayen}{\textit{University of Tübingen}}{}

\begingroup

\makeatletter
\long\def\@makefntext#1{\parindent 0pt\noindent #1}
\makeatother

\footnotetext{\textbf{Acknowledgment} The authors thank Yu-Hsiang Tseng for identifying word sense for the dataset in the current paper. This work was supported by the European Research Council under Grant SUBLIMINAL (\#101054902) awarded to R. Harald Baayen.}

\footnotetext{Yuxin Lu（卢语欣）[yuxin.lu@uni-tuebingen.de]; Department of General and Computational Linguistics, University of Tübingen, Keplerstraße 2, 72074, Tübingen, Germany. \\ 
Orcid id: \url{https://orcid.org/0009-0002-0010-7919}}
\endgroup

\englishabstract{
\noindent
In Standard Chinese, Tone 3 (the dipping tone) becomes Tone 2 (rising tone) when followed by another Tone 3.  Previous studies have noted that this sandhi process may be incomplete, in the sense that the assimilated Tone 3 is still distinct from a true Tone 2.  While Mandarin Tone 3 sandhi is widely studied using carefully controlled laboratory speech \parencite{xu_contextual_1997} and more formal registers of Beijing Mandarin \parencite{yuan_3rd_2014}, less is known about its realization in spontaneous speech,  and about the effect of contextual factors on tonal realization.  The present study investigates the pitch contours of two-character words with T2-T3 and T3-T3 tone patterns in spontaneous Taiwan Mandarin conversations. Our analysis makes use of the Generative Additive Mixed Model \parencite[GAMM,][]{wood_generalized_2017} to examine  fundamental frequency (F0) contours as a function of normalized time. We consider various factors known to influence pitch contours, including gender, {\color{black}duration}, word position, bigram probability, neighboring tones, speaker, and also novel predictors, word and word sense \parencite{Chuang:Bell:Tseng:Baayen:2025}. Our analyses revealed that in spontaneous Taiwan Mandarin, T3-T3 words become indistinguishable from T2-T3 words, indicating complete sandhi, once the strong effect of word (or word sense) is taken into account. 
}
\englishkeywords{\textbf{T}one 3 sandhi \space \space \textbf{T}onal assimilation \space \space \textbf{G}AMM \space \space \textbf{W}ord-specific tonal realization}

\vspace{1cm}

\noindent
There is an increasing interest in spontaneous speech in phonetics, psycholinguistics, and related fields. While carefully controlled speech from laboratory experiments can provide valuable insights into specific aspects of speech production, spontaneous speech allows researchers to study authentic language use in everyday conversations. Spontaneous speech differs from laboratory speech in various aspects, such as the choice of words, syntactic patterns, tones and intonation, and phonological assimilation \parencite{tucker2016why}. For instance, the pronunciation of Mandarin tones in spontaneous speech in everyday use differs from how it is described in textbooks and second-language learning classrooms, posing a huge challenge for second language learners of Mandarin.

Mandarin Chinese is a tonal language, with each syllable being distinguished by a specific lexical tone in addition to vowels and consonants. This tonal differentiation contributes to distinguishing between words' meanings. 
The main acoustic correlate of Mandarin tones is the shape of words' fundamental frequency (F0) contours. Mandarin Chinese has four lexical tones: a high-level tone (T1), a rising tone (T2), a dipping tone (T3), and a falling tone (T4), along with a neutral or floating tone \parencite{chao_grammar_1968}. \textcite{chao_grammar_1968} describes the neutral tone as being unstressed, noticeably weaker in intensity, and shorter in duration. When a syllable is placed in a disyllabic word or connected speech, tone sandhi takes place. Tone sandhi refers to the modulation of a given syllable's lexical tones by the tones of other syllables in its context. Tone 3 sandhi is the most widely studied tone sandhi phenomenon in Mandarin Chinese. In a disyllabic word with the tone pattern T3-T3, the first T3 is typically pronounced as T2, leading to the perception that the T3-T3 tone pattern sounds similar to the T2-T3 tone pattern. There has been a long-standing debate about whether this tone sandhi results in incomplete or complete assimilation of initial T3 to an initial T2 \parencite{yuan_3rd_2014}. 

Although Tone 3 sandhi has long been studied for laboratory speech \parencite{shih_prosodic_1986, xu_contextual_1997}, its realization in spontaneous speech has been less extensively investigated. It has been observed that in general, several standard acoustic cues for individual tones, such as F0, are greatly influenced by contextual factors or even reduced to a large degree in spontaneous speech, rendering them unreliable \parencite{brenner_acoustics_2013}. The majority of research in Mandarin Tone 3 sandhi is based on words that are articulated in isolation or words embedded in simple sentences or careful speech elicited in the lab using experimental tasks \parencite{warner_reduction_2011, wagner_defense_2015}. In order to gain a more comprehensive understanding of the production of Tone 3 sandhi, it is essential to examine it in the speech style that speakers and listeners most frequently and commonly use in everyday settings. The study of \textcite{yuan_3rd_2014}, addresses the realization of Tone 3 sandhi as it is realized in spontaneous spoken Mandarin, using corpora of standard Mandarin telephone speech and Mandarin broadcast news speech.

The present study {\color{black}goes a step further, and investigates} the realization of disyllabic words with T2-T3 and T3-T3 tone patterns in spontaneous spoken Taiwan Mandarin in face-to-face conversations. In what follows, section~1 provides further details on what is currently known about Tone 3 sandhi. Section~2 introduces our materials and the methods we used for data analysis. Results are presented in section 3. Our study concludes with a summary and discussion.

\section{\MakeUppercase{Incomplete neutralization of Tone 3 sandhi}}

\noindent
Most studies have reported incomplete neutralization for Tone 3 sandhi in Mandarin Chinese. The Sandhi Rising tone (SR tone, the first T3 in T3-T3) is found to exhibit lower pitch compared to the Lexical Rising tone (LR tone, T2 in T2-T3), with mean differences ranging from 3.2 Hz to 20 Hz \parencite{cui_effect_2020, kratochvil_phonetic_1984, myers_investigating_2003, shen_tonal_1990, xu_contextual_1993, xu_contextual_1997, zee_spectrographic_1980}. The SR tone has also been described as tending to have less F0 excursion in the pitch contour across various varieties of Mandarin.
For instance, \textcite{xu_contextual_1997} examined disyllabic non-word /ma-ma/ sequences with 16 possible Mandarin bi-tonal combinations embedded in carrier sentences produced by Beijing Mandarin native speakers. This study revealed that although both exhibited similar fall-rise-fall F0 contours, T3-T3 has a slightly lower pitch than T2-T3 throughout both syllables. This observation suggests for laboratory speech that the SR tone does not fully assimilate with the LR tone. However, most studies addressing Tone 3 sandhi in Taiwan Mandarin have reported that the SR tone and the LR tones are more similar to each other than in other varieties such as Beijing Mandarin. These differences in realization may remain visible but are not well supported by statistical analysis \parencite{cheng_are_2013,fon_what_1999,myers_investigating_2003, peng_lexical_2000}.

Only a few studies have examined Tone 3 sandhi in connected or spontaneous speech. An early study by \textcite{kratochvil1998intonation} analyzed the speech of a speaker of Beijing Mandarin, who was asked to generate sentences using words presented in list format.  Using discriminant analysis as classifier, he found that most of the tokens with the sandhi T3 were grouped with the lexical T3 rather than the lexical T2. This study concluded that speakers of Mandarin are adjusting the underlying T3, but the resulting SR tone is yet not equivalent to a T2. 

More recently, \textcite{yuan_3rd_2014} analyzed the realization of Tone 3 sandhi in Mandarin bi-character words in connected speech from a large corpus of telephone conversations and formal news broadcasts. They found that, despite the remarkable similarity, the LR tone displays higher and longer F0 rising than the SR tone. This low-level acoustic difference in the magnitude and the time period of F0 rise is consistent with previous results obtained using carefully controlled speech. 
Following up on the study by \textcite{xu_contextual_1997}, \textcite{wu_tone_2021} analyzed Mandarin disyllabic words with 16 tonal combinations, but used large corpora of journalistic speech. They found that the SR tone in the T3-T3 sequence is often realized as other tones, suggesting that the tone sandhi rule in connected speech is more likely to be a tendency rather than an absolute rule. In addition to differences in the speech signal, differences between the SR and the LR tone have also observed in EEG waveforms \parencite{chen_encoding_2022}. The authors argue that different stages of encoding are involved in the production of Tone 3 sandhi. 

For a variety of reasons, the way that tones are realized in connected speech is different from the canonical form. Previous research has shown that tonal context is a crucial factor in shaping tone contours \parencite{xu_contextual_1997}. When placed in context, the realization of a given tone is greatly influenced by its preceding tone and following tone, resulting in tonal co-articulation. In addition to the influence of tonal context, speaking rate and the position of a word in the utterance also play a role  \parencite{wu_tone_2021, yuan_speaking_2021}. The faster a  speaker talks, the less time is available for implementing changes in the muscles governing the vibration of the vocal cords, resulting in reduced pitch excursions \parencite{cheng_are_2013, xu_maximum_2002}. 
Furthermore, a speaker's speaking style has been found to be another factor predicting tonal variation, given that speakers has their own speaking \parencite{stanford_sociotonetics_2016}.  

The realization of tone is often also shaped by paralinguistic and sociolinguistic factors, including dialect, gender and a speaker's emotional state. An analysis of the realization of tones in several automatic speech recognition corpora, registering the speech of some 2300 speakers from seven dialectal backgrounds \parencite{tian_mandarin_2022}, reported that tone sandhi was present not only for speakers of Beijing Mandarin, but also for speakers of other dialects of Mandarin, but with regional variation in the degree of assimilation. 

In addition, it has been demonstrated that gender plays a role in tonal realizations \parencite{liang_sociophonetic_2011}. Specifically, \textcite{tian_mandarin_2022} found that male speakers tend to produce a greater F0 rise than female speakers in producing both SR tone and LR tone. A speaker's emotional state is also known to co-determine tonal realization \parencite{chang_emotional_2023, zhang_acoustic_2006}.

Furthermore, T3-T3 tone sandhi has been argued to vary also with lexical frequency \parencite{yuan_3rd_2014,tian_mandarin_2022}. The study by \textcite{yuan_3rd_2014} found that the SR tone exhibited more F0 difference from the LR tone in more frequent words than in less frequent words. Replicating and extending this study, 
\textcite{tian_mandarin_2022} also reported a similar effect of word frequency across a range of dialects.  One possible source for these frequency effects is that more frequent words tend to be realized with shorter spoken word duration, which in turn predicts reduced pitch excursion.
A random forest analysis by \textcite{wu_realization_2023} shows that, among varying factors including prosodic position, word frequency, tonal contexts, and part of speech, lexical frequency has the most contribution to the SR tone's pitch realization in naturally occurring journalist speech.


Last but not least, some research has shown that fine phonetic detail may reflect subtle differences in meaning \parencite{pierrehumbert_word-specific_2002}. More recent studies supporting this possibility have investigated the spoken word duration of English heterographic homophones \parencite{gahl_time_2024}, the duration of English syllable-final /s/ \parencite{tomaschek_phonetic_2021} and the duration of noun-verb conversion pairs \parencite{lohmann_cut_2018}.  \textcite{Chuang:Bell:Tseng:Baayen:2025} report for conversational Taiwan Mandarin that the precise shape of the T2-T4 tone pattern  is in part predictable from their meanings, while controlling for segment-related, speaker-related and contextual factors. 

In summary, incomplete neutralization, while commonly found for Beijing Mandarin, is however visible but not significant in Taiwan Mandarin. Tonal variation in general is co-determined by phonetic factors such as tonal context, speaking rate and speaker variability, by sociophonetic factors such as a speaker's sociolect and gender, by the speaker's emotional state, by frequency of use, and by the details of words' meanings.

\begin{figure}
    \centering
    \includegraphics[width=1\textwidth]{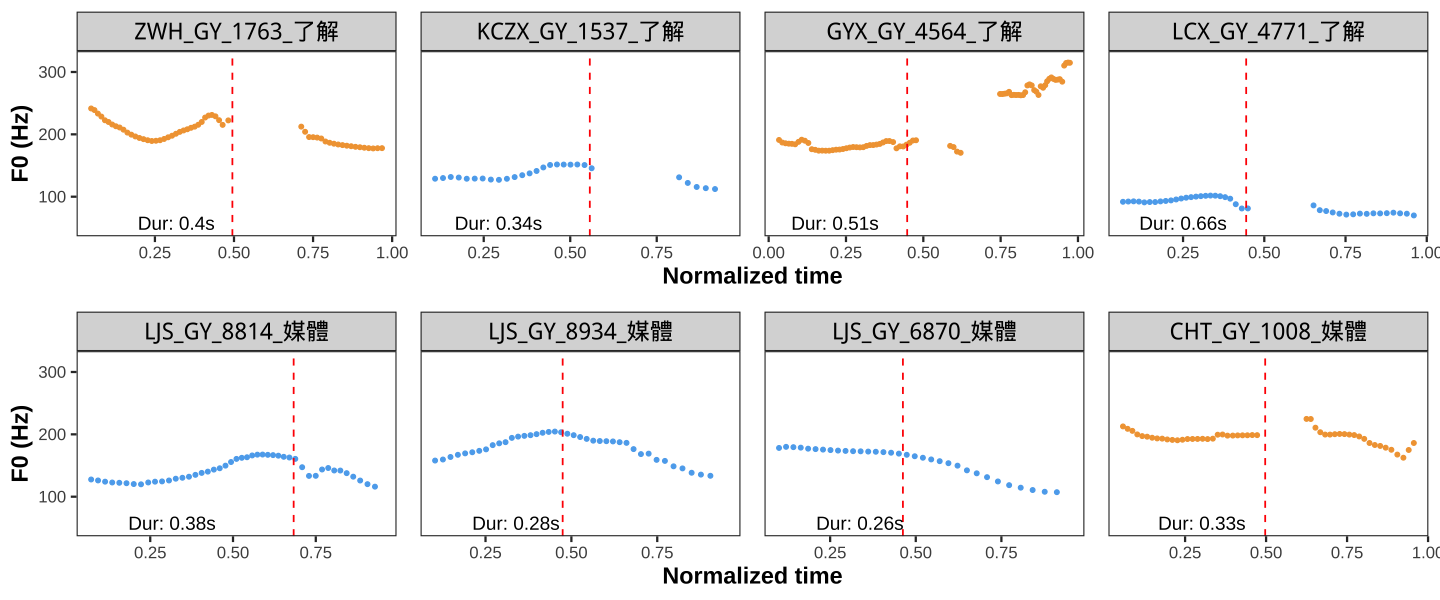}
    \caption{Example pitch contours for selected tokens of 了解\ (\textit{liao3jie3}, `to know') (upper panels) and 媒体\ (\textit{mei2ti3}, `media') (lower panels). 
    {\color{black} The pitch contours are color-coded by gender, with orange indicating female speakers and blue indicating male speakers. The vertical dashed line represents token-specific syllable boundary on the normalized time scale. The duration of each token (in seconds) is noted in the bottom of each panel.}
    }
    \label{fig:example}
\end{figure}

The present study is a follow-up of \textcite{yuan_3rd_2014}, a study that addressed the acoustic characteristics of Tone 3 sandhi in spontaneous spoken Mandarin in large corpora of broadcast news and telephone conversations. The present study, however, investigates the tonal realization of disyllabic words with T2-T3 and T3-T3 in a corpus of face-to-face conversations in Taiwan Mandarin.  Figure~\ref{fig:example} provides examples of the tone contours of two words, 了解\ (liao3jie3, \textit{to know}) (upper panels) and 媒体\ (mei2ti3, \textit{media}) (lower panels) that illustrate the immense variation that characterizes pitch contours in this corpus.  
{\color{black} \textcite{xu_contextual_1997} observed for laboratory speech that the F0 contours of T2-T3 and T3-T3 words consist of a slight fall, followed by a rise, and then a fall. A similar pattern is visible for the tokens at the left hand side (ZWH\_GY\_1763\_了解\ and LJS\_GY\_8814\_媒体), but very different realizations are found in the remaining panels. The pitch contour of ZWH\_GY\_1763\_了解, which has a relatively long duration (0.40 s), generally follows the pattern described in \textcite{xu_contextual_1997}, except that the F0 of the initial consonant [\textipa{tC}] of 解\ is not present. In contrast, LJS\_GY\_6870\_媒体\ with a short duration (0.26 s), exhibits a reduced pitch contour that appears as a falling pattern.
Besides, tonal co-articulation is another important factor of tonal variations. For example, the F0 contour of the token GYX\_GY\_4564\_了解\ is high and rising at the end, rather than expected low and falling. This is because it occurs in an utterance with exclamatory intonation 太了解啦\ (tai4liao3jie3la0, \textit{knows very well!}) and is followed by  啦\ (la0), which in this utterance is realized with relatively high pitch. The token KCZX\_GY\_1537\_了解\ exhibits an initially level pitch contour because it is preceded by 我\ (wo3, \textit{I}), whose Tone 3 is often realized as a low tone in Taiwan Mandarin \parencite{fon_what_1999, Fon2004production}.}

In our study, we leverage the power of the generalized additive model \parencite[GAM,][]{wood_generalized_2017} to come to grips with this variability and to come to a better understanding of four questions. 

First, what is the tonal realization of disyllabic words with T2-T3 and T3-T3 tone pattern in spontaneous Taiwan Mandarin speech? Is there any difference between the tonal realization of the T2-T3 tone pattern and the T3-T3 tone pattern with tone sandhi? At the outset of the research reported below, we had no hypothesis about whether T3-T3 tone sandhi in Taiwan Mandarin is incomplete or complete. Anticipating the results reported below, for the data that we investigated, T3-T3 tone sandhi appears to be complete.   

Second, how does context affect tonal realizations?  In the light of the results of the previous studies of Mandarin tones, we expect to replicate the effects of speaking rate, tonal context, the position of a word in its utterance, its conditional probability in context, the gender of the speaker, and speaker-specific habits of realising tone.

Third, does a word's frequency of occurrence co-determine the shape of its pitch contour?  Our hypothesis with respect to this question is that once word duration and speaking rate are taken into account, frequency does not further modulate pitch. This hypothesis is motivated by two considerations. First, frequency of use is a lexical property that is time-invariant, and it is unclear to us why it would modulate pitch over time. Second, if frequency of use would have a systematic effect on the realization of pitch, one would expect this modulation to be visible across all tonal patterns in the same way. No such effect has been reported in the existing literature.

Fourth, is the shape of a word's pitch contour co-determined by its meaning? Based on the results obtained by \textcite{Chuang:Bell:Tseng:Baayen:2025} for the T2-T4 tone pattern in conversational Taiwan Mandarin, we expect that the fine details of how T2-T3 and T3-T3 tone contours are realized are word-specific and semantic in nature. If indeed words' meanings co-determine their pitch contours, then this in turn raises the question of whether evidence for incomplete neutralization is robust when words and words' senses are taken into account as control variables.

{\color{black}In the current study, the statistical analysis will be conducted on entire words, including consonant onsets and vowels. For example, for 媒体\ (\textit{mei2ti3}), the analysis is performed on [\textipa{me\textsuperscript{I}t\textsuperscript{h}i}]. This approach aligns with the general framework of the Discriminative Lexicon Model \parencite{Chuang:Bell:Tseng:Baayen:2025,Heitmeier:Chuang:Baayen:2025}, which focuses on the relationship between semantics and phonetic realization. Previous work has shown that semantics, operationalized using contextualized embeddings, predicts the pitch contours of word tokens \parencite{Chuang:Bell:Tseng:Baayen:2025, lu2026realization}. Therefore, our focus is on how pitch is realized on whole onomasiological words rather than on their constituent morphemes or syllables.}

\section{\MakeUppercase{Materials and methods}}

\subsection{Data} \label{sec:data}

\noindent
Data analyzed in the present paper is taken from the Taiwanese Mandarin Spontaneous Speech Corpus \parencite{fon_preliminary_2004}, which consists of 30 hours of spontaneous conversations from speakers aged between 20 and 60 (31 females and 24 males). 
{\color{black}
The audio recordings were orthographically transcribed using traditional Chinese characters. Word and character boundaries in the orthographic representation were identified using the word segmentation system developed by Academia Sinica \parencite{ma__2001}. The transcriptions written in Chinese characters were then romanized. The audio and transcripts were forced-aligned at both the word and character level with EasyAlign \parencite{goldman2011easyalign}. The resulting alignments were manually checked and corrected by native speakers when necessary. In the current study, we followed the segmentation provided in the corpus. The Hanyu Pinyin transcriptions and canonical tones were obtained from the \texttt{pinyin\_jyutping\_sentence} package for Python. For Chinese words with multiple possible pronunciations, the tonal specifications were verified and corrected when necessary (e.g., 东西\ can be pronounced either as \textit{dong1xi0} `thing' or as \textit{dong1xi1} `east and west').

F0 values were estimated using Praat's \texttt{To Pitch (cc)} function with the time step set at 0.001 seconds and a Gaussian window for optimal accuracy \parencite{boersma_praat_2020}. 
Pitch floor was set at 75 Hz and pitch ceiling at 400 Hz for female speakers, and pitch floor at 50 Hz and pitch ceiling at 300 Hz for male speakers. No F0 values were interpolated for the voiceless sections of the tokens.
}


We extracted 3937 word tokens with the T2-T3 tone pattern, representing 299 word types, and 4791 word tokens with the T3-T3 tone pattern, representing 269 word types. To ensure that for statistical analysis, sufficient numbers of tokens are available for each word type, words with fewer than 12 tokens were excluded from our dataset. {\color{black}Following \textcite{Chuang:Bell:Tseng:Baayen:2025}, }for high-frequency words with more than 200 tokens, we randomly sampled 200 tokens, in order to avoid model predictions from being biased towards high-frequency words. {\color{black}This led to the exclusion of 1400 tokens of 7 high-frequent word types.} As a consequence, every word type has a maximum of 200 tokens and a minimum of 12 tokens. Furthermore, in order to facilitate the statistical analysis of effects of gender, words contributed by only female speakers or only male speakers were excluded. We also made sure that tokens of one word type are contributed by two or more speakers, so as to avoid model prediction being biased by one specific speaker’s way of speaking. Tokens with potential F0 tracking errors were removed on the basis of visual inspection. 
{\color{black}Table~\ref{tab:data} presents an overview of tokens and word types after data sampling. The first resulting dataset contains 1462 tokens representing 20 word types with tone pattern T2-T3 and 878 tokens representing 20 word types with tone pattern T3-T3 (see Table~\ref{tab:data} for the dataset of words).}

The study by \textcite{Chuang:Bell:Tseng:Baayen:2025} argues that semantics co-determines the tonal realizations. In support of this hypothesis, they show that prediction of pitch contours improves when (orthographic) word is replaced by word sense. Therefore, we made use of the same word sense identification system as used in their study, described in \textcite{hsieh_resolving_2024}, which uses BERT in combination with the Chinese WordNet \parencite{huang_chinese_2010}. 

In our dataset, there are four words which have no word sense in the Chinese WordNet. For the remaining 36 words in our dataset, a total of 71 senses was identified. Most of the words have 1 to 3 senses.  The distribution of  senses is skewed towards the right, with about one-third of the senses having 6 tokens or fewer. To make sure that all senses have sufficient tokens for statistical evaluation, we first removed senses with fewer than 8 tokens, {\color{black}which resulted in the exclusion of one word type}. In addition, to prevent model predictions from being biased in favor senses with a large number of tokens, we sampled 40 tokens for each of the more frequently used senses. The second resulting dataset, prepared specifically for assessing the role of sense as opposed to orthographic word, contains in total 1144 tokens of 46 sense types, corresponding to 35 orthographic word types {\color{black}(see the sense dataset in Table~\ref{tab:data})}.

\begin{table}[htbp]
    \centering
    \tablesize
    \caption{Summary of tokens and word types in each tone pattern in the word and sense analyses.}
    \label{tab:data}
    \begin{tabular}{lccccc}
        \toprule
        & \multicolumn{2}{c}{\textbf{Word dataset}} & \multicolumn{3}{c}{\textbf{Sense dataset}} \\
        \cmidrule(lr){2-3} \cmidrule(lr){4-6}
        \textbf{Tone pattern} & Tokens & Word types & Tokens & Word types & Sense types \\
        \midrule
        T2-T3 & 1462 & 20 & 567 & 16 & 20 \\
        T3-T3 & 878 & 20 & 577 & 19 & 26 \\
        \textbf{Total} & \textbf{2340} & \textbf{40} & \textbf{1144} & \textbf{35} & \textbf{46} \\
        \bottomrule
    \end{tabular}
\end{table}

\subsection{Predictors}

\noindent
In our statistical analyses, the response variable is log-transformed F0. The log-transformation was motivated by a Box-Cox analysis \parencite{Box:Cox:1964}. We modelled log-transformed pitch as a function of normalized time. 

{\color{black}Time is an important factor co-determining the realization of pitch contours, which is why token duration and its interaction with normalized time are included in the GAMM models. In the current study, however, we are interested in the shape of pitch contours, for which working in normalized time is essential. For example, consider a sine wave with some fixed amplitude, but one with a period of $2\pi$ and the other with a period of $4\pi$. Over the interval $[0,2\pi]$, the curves appear dissimilar, but when normalized to the [0,1], it is clear that their shapes are identical. Since a word's duration can be predicted using a separate model \parencite[see, e.g.,][]{gahl_time_2024}, shape and duration models jointly can account for observed pitch contours in actual real time.} 

{\color{black}Time normalization was carried out on entire words, including onset consonants\footnote{
{\color{black}
For detailed discussion of the way in which GAMs deal in a principled way with voiceless initial or medial obstruents, see \textcite{Chuang:Bell:Tseng:Baayen:2025}.}
}and final nasals if present.}
Table \ref{tab:predictors} provides an overview of variables that we considered in the statistical analysis.

\begin{sidewaystable}
\centering
\tablesize
\caption{Predictors used in the statistical analysis}
\label{tab:predictors}
\renewcommand{\arraystretch}{1.00}  
\begin{tabular}{p{6cm} p{15cm}}
\toprule
\textbf{Variable (Abbreviation in R)} & \textbf{Definition} \\
\midrule
Log-transformed pitch (\texttt{logF0}) & Natural logarithm (log$_e$) of the fundamental frequency (F0). \\

Normalized time (\texttt{normalized\_t}) & For each token, time was normalized to a range between 0 and 1, allowing tokens with varying durations to be modeled on a common scale. \\

Gender (\texttt{gender}) & A factor of two levels: \textit{male} and \textit{female}. \\

Tone pattern (\texttt{tone\_pattern}) & The tonal pattern of the token, represented as a factor of two levels: \texttt{23} (rising-dipping) and \texttt{33} (dipping-dipping). \\

GenderXtone (\texttt{genderXtone}) & A factor for the interaction between \texttt{gender} and \texttt{tone\_pattern}, with four levels: \textit{female.23}, \textit{male.23}, \textit{female.33}, and \textit{male.33}. This factor is used to account for gender-related differences in tone realization. \\

{\color{black}Duration (\texttt{duration})} & {\color{black}The duration of the token in seconds. \texttt{Duration} was included as a covariate to control for the effect of time in tonal realization.} \\

Word position in utterance

(\texttt{norm\_utt\_pos})
& The normalized position of a token within its utterance, ranging from 0 to 1. \textcolor{black}{Smaller values indicate earlier positions, and larger values indicate later positions in the utterance.} Words in single-word utterances were coded as 1. \\

Bigram probability of the preceding word (\texttt{bg\_prob\_prev}) & Probability of the target word given the preceding word (following \textcite{gahl_time_2008}), measuring how predictable the word is in context. Higher values indicate greater predictability.\\

Bigram probability of the following word (\texttt{bg\_prob\_fol}) & Probability of the following word given the target word. \\

Tonal context (\texttt{tonal\_context}) & The interaction of the preceding and following tones of a token. Each tone (or a pause) can take six values: \texttt{0, 1, 2, 3, 4, or PAUSE}, resulting in 36 possible tonal contexts.\\

Speaker (\texttt{speaker}) & A factor with 55 anonymized speaker identifiers. \\

Word (\texttt{word}) & A factor with 40 orthographic word forms. All words in this study are two-character words (e.g., 媒体\ (\textit{mei2ti3}, `media'). \\

Sense (\texttt{sense\_type}) & A factor with 46 levels representing word senses rather than orthographic forms. \\

\bottomrule
\end{tabular}
\end{sidewaystable}

\subsection{Statistical Analysis}

\noindent
One of the major challenges in analyzing dynamic speech data, such as pitch contours, formant trajectories and tongue movements, is non-linear change over time  \parencite{wieling_investigating_2016, wieling_analyzing_2018,chuang_analyzing_2021}. The Generalized Additive Mixed Model \parencite[GAMM,][]{wood_generalized_2017}  has emerged as a flexible and powerful statistical method for analysing non-linear time-series data. It has been successfully applied to the analysis of F0 contours in various languages, including Mandarin \parencite{Chuang:Bell:Tseng:Baayen:2025}, Brazilian Portuguese \parencite{da_silva_miranda_visual_2020},  Papuan Malay \parencite{kaland_red_2023}, and French \parencite{deng_pitch_2023}.

In the present study, we make use of the \textbf{mgcv} package \parencite{wood_generalized_2017} for R \parencite{team_r_2020} , using the \texttt{bam} function to model pitch as a function of time and the predictors listed above.  
{\color{black}To account for autocorrelation in the residuals, we include a first-order autoregressive correlation (AR(1)) in the model. This assumes that the residual at time $t$ is correlated with the residual at time $t-1$, allowing the model to appropriately handle temporal dependence in the pitch contours. An AR(1) with $\rho = 0.89$ removes most of the autocorrelation.

We use thin plate regression spline smooths to model the effect of time and its interactions with gender and tone pattern. These splines allow the model to flexibly capture how pitch contours unfold over time across groups and are well suited for continuous, non-linear trajectories such as fundamental frequency (F0) contours. For an introduction to the use of GAMMs in phonetics, see \textcite{wieling_analyzing_2018}; for mathematical details, see \textcite{wood_generalized_2017}. 

Thin plate regression splines are also applied to continuous predictors such as duration, word position, and bigram probabilities. Tensor product smooths are used to model interactions between time and continuous predictors, with the \texttt{ti()} directive to isolate the interaction effect from its corresponding main effects. Factor smooths for time are incorporated for tonal context, speaker, and word.} These factor smooths (specified as \texttt{s(time, predictor, bs="fs", m=1)}) represent nonlinear random effects of subject and word, and include shrunk estimates of by-subject and by-word intercepts.

\section{\MakeUppercase{Results}} \label{sec:results}

\noindent
The model that we propose for the present data was obtained by fitting a series of increasingly complex GAMM to the dataset. The best-fit model resulting from this exploratory investigation of the Mandarin pitch contours is specified follows:

{\color{black}
\begin{tabbing}
mmmmm\=mm\= \kill
\texttt{logF0}  \> $\sim$ \> \textbf{\texttt{genderXtone +}} \\
      \> \> \textbf{\texttt{s(normalized\_t, by=genderXtone) +}} \\
      \> \> \texttt{s(duration, k=5) +} \\
      \> \> \texttt{ti(normalized\_t, duration) +} \\
      \> \> \texttt{s(normalized\_utt\_pos, k=6) +} \\
      \> \> \texttt{ti(normalized\_t, normalized\_utt\_pos) +} \\
       \> \> \texttt{s(bg\_prob\_prev, k=5) +} \\
       \> \> \texttt{ti(normalized\_t, bg\_prob\_prev) +} \\
       \> \> \texttt{s(bg\_prob\_fol, k=5) +} \\
       \> \> \texttt{ti(normalized\_t, bg\_prob\_fol) +} \\
        \> \> \texttt{s(normalized\_t, tonal\_context, bs=`fs', m=1) +} \\
       \> \> \texttt{s(normalized\_t, speaker, bs=`fs', m=1) +} \\
       \> \> \textbf{\texttt{s(normalized\_t, word, bs=`fs', m=1)}} \\
\end{tabbing}
}

Table~\ref{tab:AIC} therefore reports the increase in Akaike Information Criteria (AIC) when conceptually related sets of terms in the model are withheld from the GAMM.  Invariably, these increases are large, indicating that all terms are contributing to making model predictions more precise. Table \ref{tab:gam.summary} provides a summary of this model. 
In what follows, we first briefly discuss the parametric coefficients of this model, and then report in more detail on the smooth terms of the model.

\begin{table}[htbp]
\centering
\tablesize
\caption{Increase in AIC when one or more conceptually related predictors are removed from the best-fit GAMM.}
\begin{tabular}{lr}
\toprule
\textbf{Model term} & \textbf{$\Delta$AIC} \\ 
\midrule
Speaker & 9781.80 \\ 
Word & 4242.28 \\ 
Tonal context & 1582.08 \\ 
Duration & 832.87 \\ 
Word position in utterance & 579.02 \\ 
Bigram probability of preceding word & 169.92 \\ 
Bigram probability of following word & 66.93 \\ 
GenderXtone & 26.87 \\ 
\bottomrule
\end{tabular}
\label{tab:AIC}
\end{table}

\subsection{Parametric Coefficients}

\noindent
The reference level of the \texttt{genderXtone}  interaction is  the T2-T3 tone pattern for female speakers.  The contrast effect for T2-T3 for male speakers indicates that, as expected, the male speakers' T2-T3 contour is realized with a lower overall pitch. The T3-T3 intercept for female speakers does not differ from the general intercept, and the T3-T3 contrast for male speakers is again pointing to a lower overall pitch.  Given that the two contrasts for male speakers are very similar {\color{black}(-0.5051 and -0.5328)},  the parametric coefficients considered jointly simply indicate that, irrespective of tone pattern, male speakers realized pitch contours with lower voices.

\begin{table}[htbp]
\centering
\tablesize
\caption{Model summary of GAMM best fitted to the pitch contours of Taiwan Mandarin T2-T3 and T3-T3 words.} 
\begin{tabular}{lrrrr}
   \hline
A. parametric coefficients & Estimate & Std. Error & t-value & p-value \\ 
  (Intercept) & 5.2730 & 0.0261 & 201.8577 & $<$ 0.0001 \\ 
  genderXtonemale.23 & -0.5051 & 0.0319 & -15.8256 & $<$ 0.0001 \\ 
  genderXtonefemale.33 & -0.0226 & 0.0214 & -1.0588 & 0.2897 \\ 
  genderXtonemale.33 & -0.5328 & 0.0383 & -13.9073 & $<$ 0.0001 \\ 
   \hline
B. smooth terms & edf & Ref.df & F-value & p-value \\ 
  s(normalized\_t):genderXtonefemale.23 & 4.9301 & 5.2039 & 8.6285 & $<$ 0.0001 \\ 
  s(normalized\_t):genderXtonemale.23 & 1.2680 & 1.3176 & 8.6629 & 0.0012 \\ 
  s(normalized\_t):genderXtonefemale.33 & 4.4111 & 4.7620 & 5.3698 & 0.0001 \\ 
  s(normalized\_t):genderXtonemale.33 & 2.0540 & 2.2383 & 1.7614 & 0.1596 \\ 
  s(duration) & 3.4681 & 3.8449 & 41.1607 & $<$ 0.0001 \\ 
  ti(normalized\_t,duration) & 21.8894 & 23.8699 & 29.4859 & $<$ 0.0001 \\ 
  s(normalized\_utt\_pos) & 1.0013 & 1.0024 & 532.0097 & $<$ 0.0001 \\ 
  ti(normalized\_t,normalized\_utt\_pos) & 20.3819 & 23.1806 & 4.6625 & $<$ 0.0001 \\ 
  s(bg\_prob\_prev) & 2.9280 & 3.5840 & 43.8531 & $<$ 0.0001 \\ 
  ti(normalized\_t,bg\_prob\_prev) & 12.2691 & 15.9500 & 2.6771 & 0.0003 \\ 
  s(bg\_prob\_fol) & 3.2000 & 3.8685 & 1.7859 & 0.2271 \\ 
  ti(normalized\_t,bg\_prob\_fol) & 17.6078 & 20.8375 & 5.0712 & $<$ 0.0001 \\ 
  s(normalized\_t,tonal\_context) & 255.7701 & 324.0000 & 5.9977 & $<$ 0.0001 \\ 
  s(normalized\_t,speaker) & 422.0775 & 494.0000 & 21.9715 & $<$ 0.0001 \\ 
  s(normalized\_t,word) & 305.0403 & 359.0000 & 13.4559 & $<$ 0.0001 \\ 
   \hline
\end{tabular}
\label{tab:gam.summary}
\end{table}

\subsection{Smooth Terms} 

\subsubsection{Time by gender by tone pattern}

\begin{figure}[htbp]
    \centering
    \begin{subfigure}[b]{\linewidth}
        \centering
        \includegraphics[width=\linewidth]{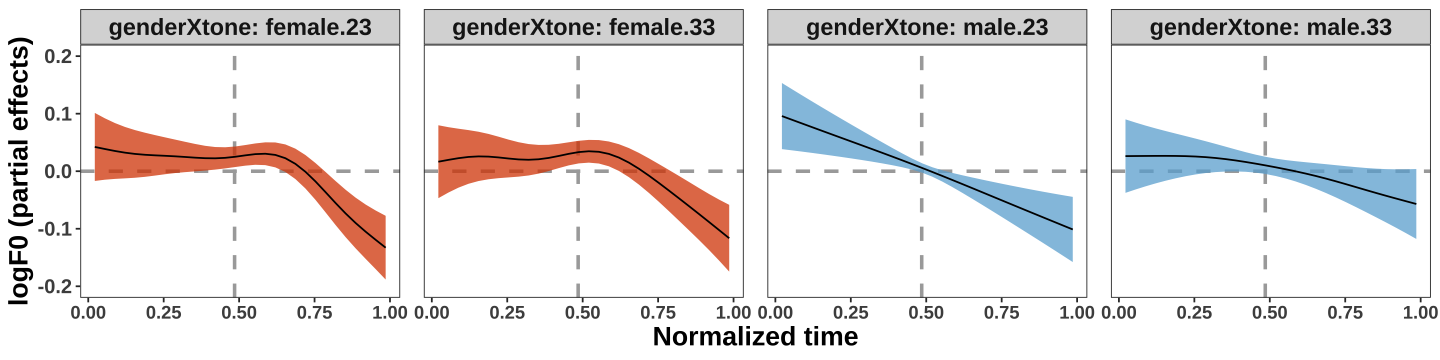}
    \end{subfigure}
    \hfill
    \begin{subfigure}[b]{\linewidth}
        \centering
        \includegraphics[width=\linewidth]{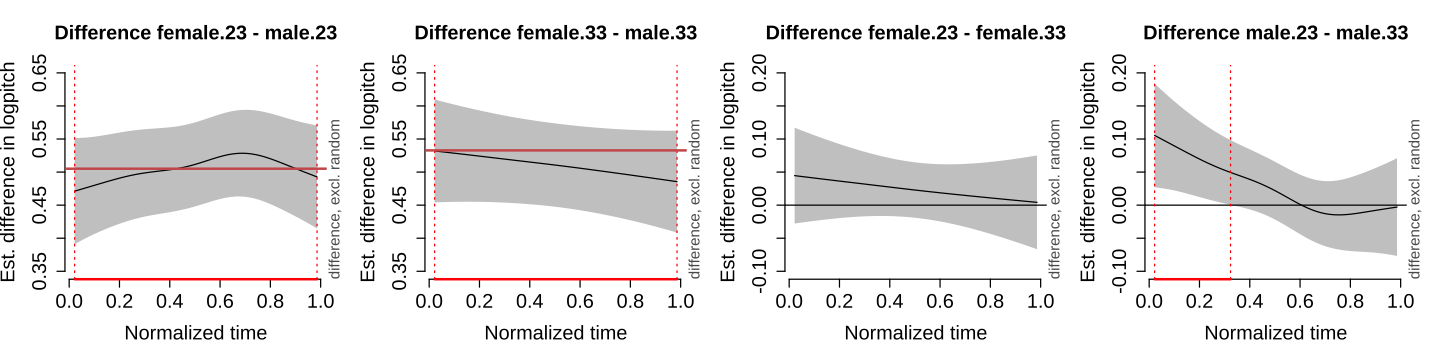}
    \end{subfigure}
    \caption{{\color{black}The partial effects of the three-way interaction of time by gender by tone pattern. The two upper left panels show the T2-T3 and T3-T3 partial effects for females, and two upper right panels represent the partial effect of tone pattern T3-T3 for male speakers respectively.
    The grey red vertical line indicates the average syllable boundary in normalized time scale (x=0.485). The grey red horizontal line indicates the y=0 reference line.
    The lower left two panels show the differences between the tonal contours for females and males for T2-T3 and T3-T3 respectively. The lower right two panels present the differences within gender for T2-T3 and T3-T3, for females and males respectively.
    The lower four panels show the estimated difference F0 trajectories between gender groups and tone pattern groups. The areas where the difference is significantly different from 0 are marked in red.}
    }
    \label{fig:genderXtone}
\end{figure}

\noindent
The first non-linear term in the model, \texttt{s(normalized\_t, by=genderXtone)} requests four smooths of time, one for each combination of \texttt{gender} and \texttt{tone pattern}. {\color{black}The smooth term for \texttt{female.23}, \texttt{male.23}, and \texttt{female.33} are significant, but the smooth term for \texttt{male.33} is not significant. The upper panels of Figure \ref{fig:genderXtone} illustrates the partial effect of each gender within the two tone patterns. The contours for \texttt{male.23} and \texttt{male.33} are almost linear. The panels for T2-T3 and T3-T3 produced by female speakers (shown in orange) appear fairly similar. However, for male speakers, the T2-T3 effect has a steeper decline than the T3-T3 partial effect (shown in blue).}  

{\color{black}To assess the statistical significance of F0 differences between groups, difference curves were computed (lower panel of Figure~\ref{fig:genderXtone}). These curves compare trajectories across gender and tone pattern groups, highlighting regions where the difference is significantly different from zero (marked in red). As expected, the comparison between male and female speakers shows a consistent difference across the entire normalized time interval. This reflects a difference in overall pitch height because female speakers in general have higher F0 than males, but not in pitch contour, as indicated by the horizontal line representing the average difference remaining entirely within the gray non-significance band. Among female speakers, no significant difference is observed between the T2-T3 and T3-T3 patterns. In contrast, male speakers show a significant difference between the two tone patterns between 0 and 0.3 on the normalized time scale. Using the male intercept logF0 (5.2370), a difference of 0.1 logF0 units corresponds to a change from approximately 125.45 Hz to 113.51 Hz, which is about an 11.94 Hz decrease. This difference is well-noticeable at the beginning but quickly diminishes to near zero by around 0.33 normalized time units.} 

{\color{black}To further investigate this three-way interaction, Table~\ref{tab:aic_word} presents four models with different model structures regarding tone pattern and gender. The interaction of normalized time by gender by tone pattern is supported by a lower AIC compared to a model that only includes a general smooth for \texttt{normalized time} ($\Delta$AIC: 28.31), and the model that includes a by-gender smooth for \texttt{normalized time} as well ($\Delta$AIC: 27.54). However, model comparison based on the fREML scores (evaluated using the \texttt{compareML} function from the \textbf{itsadug} package) suggests that there is only small difference in fREML (model with by-\texttt{toneXgender} smooth: -138107.9 and model with by-\texttt{gender} smooth: -138112.9).} 

{\color{black}In summary, the three-way interaction is not receiving unambiguous support, so the effect observed for men only may not be replicable. We therefore conclude, first, that female and male speakers realize the T2-T3 and T3-T3 tones somewhat differently, and second,  that the T2-T3 and T3-T3 patterns have the same realization for females, and probably also for males.}

\begin{table}[htbp]
\centering
\tablesize
\caption{Four models including all other predictors but with different structures regarding \texttt{tone pattern} and \texttt{gender}.}
\begin{tabular}{lr}
\toprule
 \textbf{Model} & \textbf{AIC} \\ 
\midrule
  \textbf{All other predictors} &  \\
  \hspace{1em}+ a general smooth for \texttt{normalized time}  & -278266.05 \\ 
  \hspace{1em}+ by-\texttt{tone\_pattern} smooth & -278265.93 \\ 
  \hspace{1em}+ by-\texttt{gender} smooth  & -278266.82 \\ 
  \hspace{1em}+ by-\texttt{toneXgender} smooth & -278294.36  \\ 
\bottomrule
\end{tabular}
\label{tab:aic_word}
\end{table}

\begin{figure}[htbp]
    \centering
    \includegraphics[width=\linewidth]{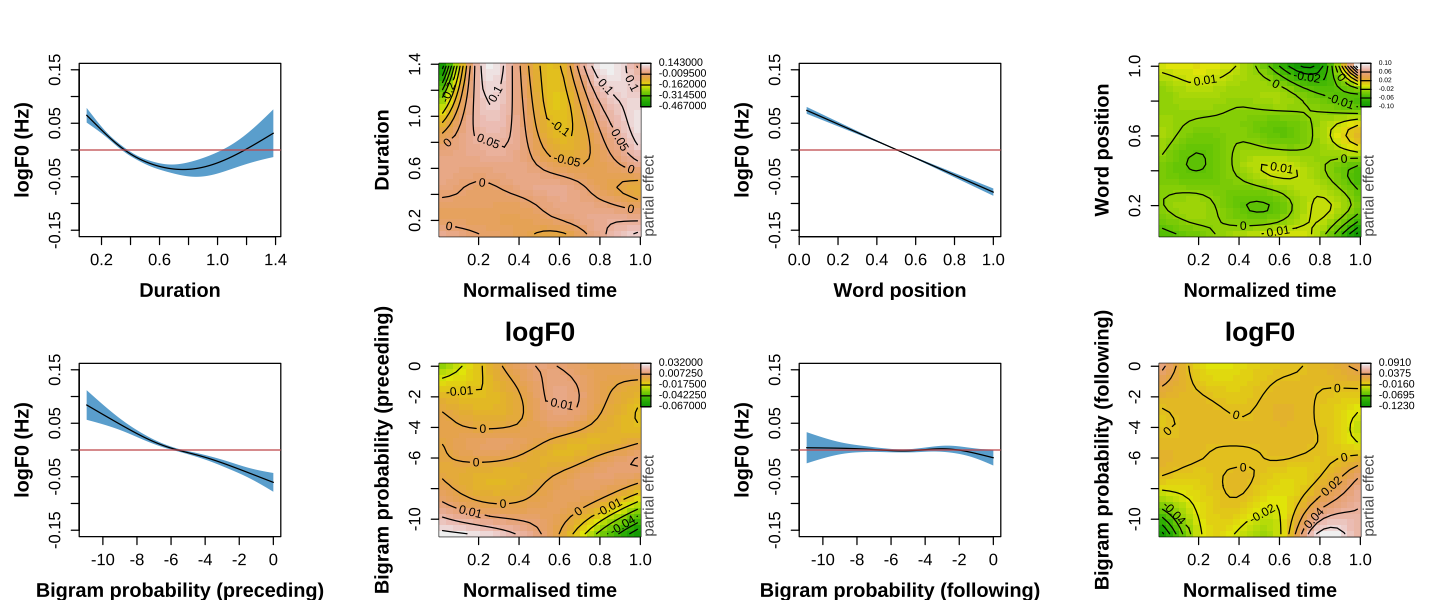}
    \caption{The partial effect of the smooth term and the interaction between normalized time and predictor. duration, word position, bigram probability of preceding word, and bigram probability of following word.}
    \label{fig:continuous}
\end{figure}

{\color{black}
\subsubsection{Duration by time}

\noindent
The upper-left panels of Figure~\ref{fig:continuous} visualizes the interaction of duration by time, which we modeled with a thin plate regression smooth for the main effect of duration and a tensor product smooth for the interaction of time by duration. Short duration indicates higher pitch, and pitch decreases as the duration becomes longer. Very longer duration show wider confidence interval due to data sparsity. In panel 2, tokens of longer duration show a small undulating pattern that may be due to to only three word tokens with exceptionally long durations. We therefore refrain from further theoretical interpretation of this partial effect. \footnote{\color{black}{There is evidence that the main effects of the duration, and bigram probabilities are further modulated by gender, but in order to avoid further model complexity, these interactions were not included.  We note here that inclusion of these interactions further improves model fit, but does not affect the partial effects that we report and discuss, and hence are only worked out in the supplementary materials.}}
}

\subsubsection{Time by word position}

\noindent
Words that occur later in an utterance tend to have lower pitch (see panel~3 of Figure \ref{fig:continuous}), as expected \parencite{shih2000declination}. The contour plot for the interaction of word position by time in this figure is less straightforwardly interpretable, and the very large effects near the end of the word are likely unreliable, given that words in utterance-final position and single-word utterances were both given 1.0 as word position.

{\color{black}
\subsubsection{Time by probability}

\noindent
The lower panels of Figure~\ref{fig:continuous} presents the partial effects involving the probabilities of the preceding and following words.  The leftmost panel presents the main effect of the bigram probability of the preceding word: as the bigram probability increases, log F0 decreases. This large main effect is modulated by a small interaction with time (second panel). The third panel clarifies that there is no main effect of the bigram probability of the following character. The fourth panel shows that there is a small interaction of this probability with time, especially for words with low probabilities. We are not aware of theoretical considerations that would predict the tensor product interaction surfaces. For the current research, we leave these surfaces in the model to optimally control for effects of preceding and following bigram probability.}

\subsubsection{Tonal context}

\begin{figure}[htbp]
    \centering
    \includegraphics[width=0.75\textwidth]{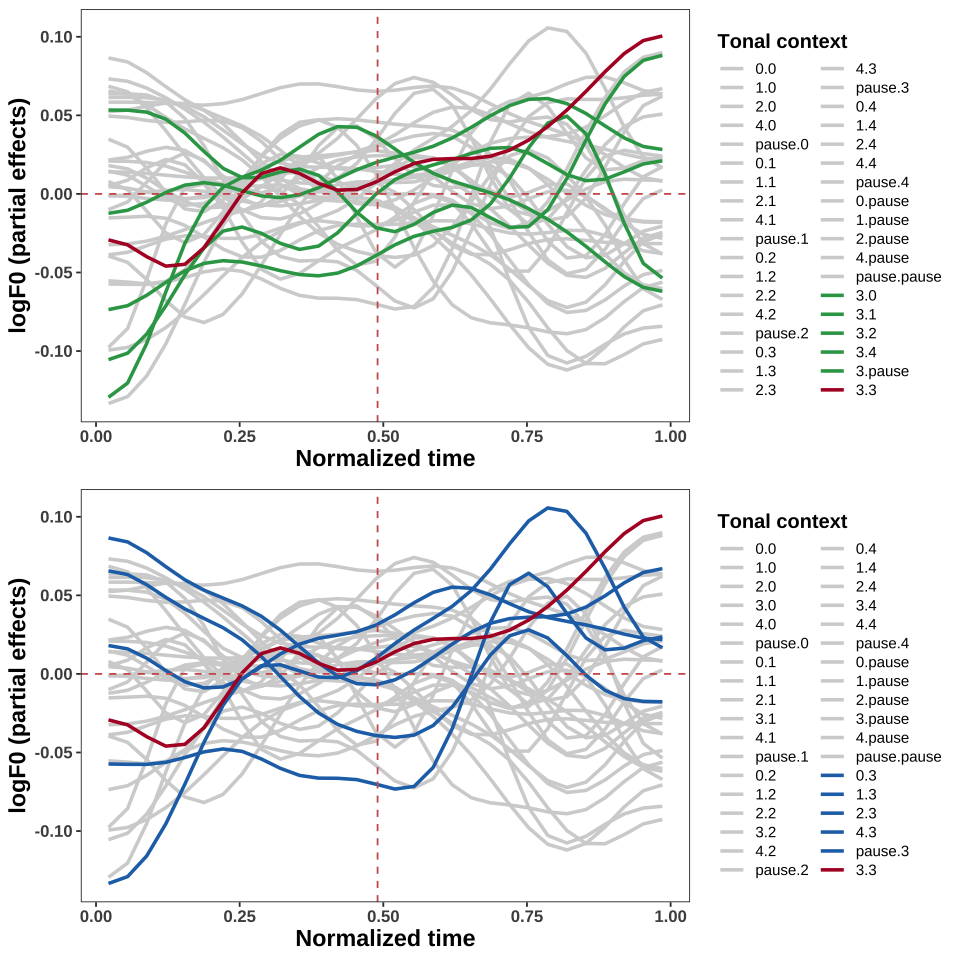}
    \caption{{\color{black}
    Partial effect of tonal context, showing all 36 tonal context levels grouped by color for visualization in the upper and lower panels. In both panels, the partial effect of the pitch contour for the 3.3 context is shown in red. In the upper panel, tonal contexts with a preceding T3 (3.1, 3.2, 3.4, and 3.pause) are highlighted in green, while all other remaining contexts are shown in grey. In the lower panel, tonal contexts with a following T3 (0.3, 1.3, 2.3, 4.3, and pause.3) are highlighted in blue. Again, all remaining other contexts are shown in grey.
    }}
    \label{tonal_context}
\end{figure}

\noindent
Figure~\ref{tonal_context} shows how the pitch contour varies with the 36 levels of tonal context, while keeping all other predictors constant. {\color{black}In both panels, the same partial effects are displayed. The partial effect of the pitch contour for the 3.3 context is shown in red. In the upper panel, tonal contexts with a preceding T3 (3.1, 3.2, 3.4, and 3.pause) are highlighted in green, while all remaining contexts are shown in grey. In the lower panel, tonal contexts with a following T3 (0.3, 1.3, 2.3, 4.3, and pause.3) are highlighted in blue, with all other contexts again shown in grey. By way of example, consider the red} curve that starts in the lower left and ends in the upper right. This curve represents words embedded between words ending and beginning with a dipping tone (Tone 3) (e.g., in our data, 你可以解释\ [\textit{ni3\ ke3yi3\ jie3shi4}, `you can explain']). These context-specific pitch contours show great variability. Apparently, the preceding dipping tone has a lowering effect on pitch {\color{black}(see upper panel)}, while the following dipping tone has a raising effect {\color{black}(see lower panel)}. {\color{black} This might be due to Tone 3 often being realized as a low-falling tone in Taiwan Mandarin \parencite{fon_what_1999, Fon2004production}} \footnote{{\color{black}There are 375 tokens with either X.T2.T3.T3 or X.T3.T3.T3 tone sequences. Tone sandhi becomes more complex when it is followed by a T3 across word boundaries. Additionally, the final T3 may be realized as a T2 if it is followed by another T3, which occurs in 9 tokens in our dataset. GAMMs fitted to the dataset both with and without these 375 tokens yield quantitatively and qualitatively very similar results.}}. 

\subsubsection{Speaker}

\begin{figure}[htbp]
    \centering
    \includegraphics[width=0.75\textwidth]{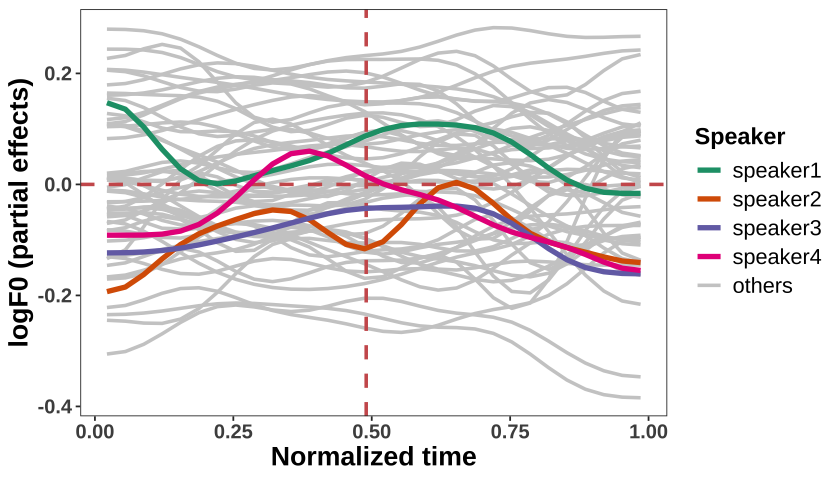}
    \caption{\color{black} Partial effect plots for speaker-specific modulations of the pitch contour, estimated by the factor smooth for speaker. There are 55 speakers in total; four selected speakers are highlighted in distinct colors, while the remaining speakers are shown in grey. Speakers 1 and 2 are female, whereas speakers 3 and 4 are male. The red dashed vertical line represents the average syllable boundary of disyllabic words in the normalized time scale. The red dashed horizontal line indicates the y=0 reference line.}
    \label{fig:speaker}
\end{figure}

\noindent
In order to take the structural variability associated with individual speakers into account, we included by-speaker factor smooths for time. Figure~\ref{fig:speaker} presents four speaker-specific pitch ``habits", for two females ({\color{black}speaker~1 and speaker~2}) and two males ({\color{black}speaker~3 and speaker~4}). These modulations of the pitch contour are independent of the gender-specific effects reported above (see Figure~\ref{fig:genderXtone}). Speaker 1 has a higher pitch than speaker 2, and within the set of male speakers, speaker 4 has a higher pitch than speaker 3.  Each of these four speakers further modulate their pitch contours in their own way.


\subsubsection{Word}

\begin{figure}[htbp]
    \centering
    \includegraphics[width=\textwidth]{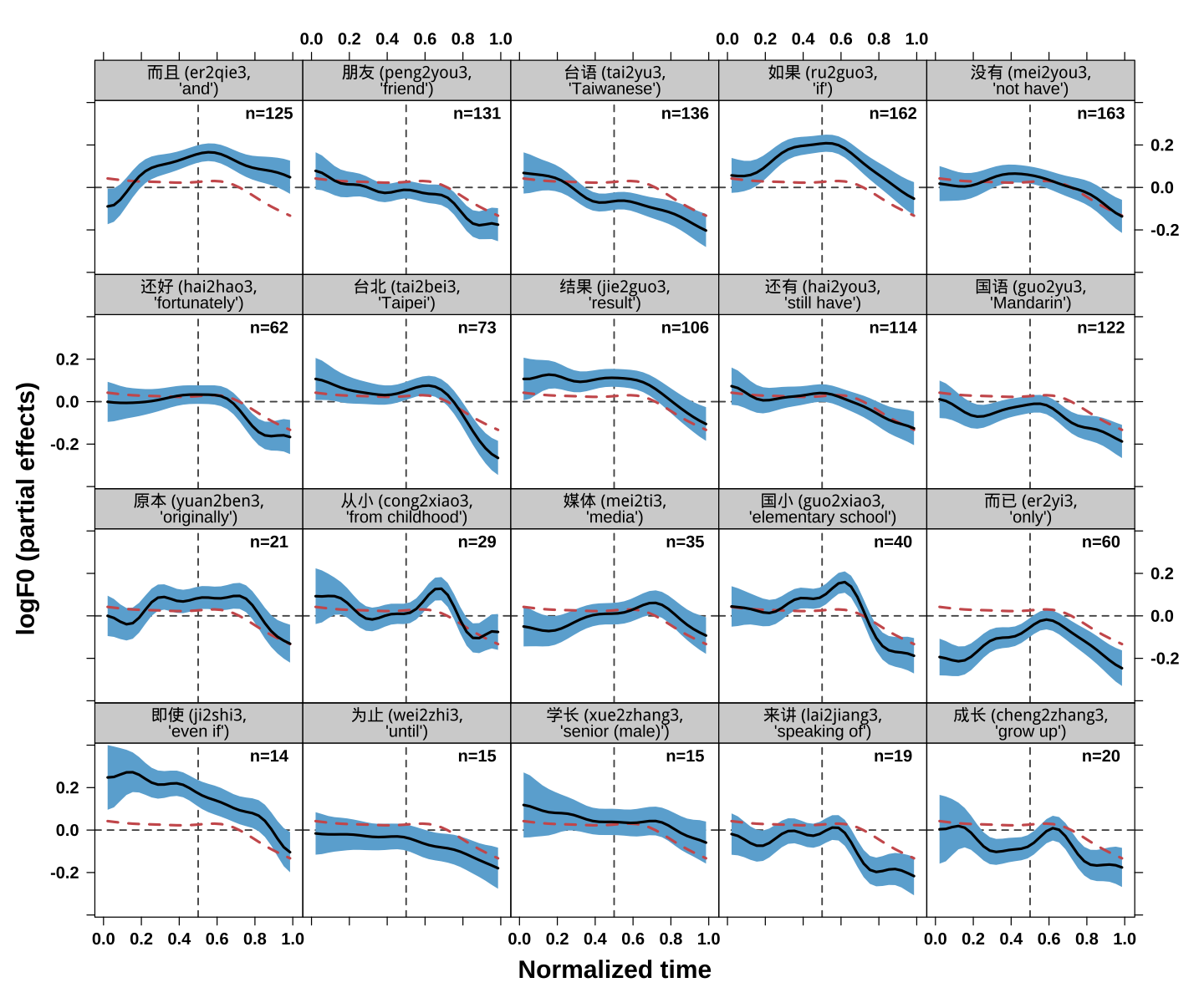}
    \caption{{\color{black}Predicted pitch contours for words with the T2–T3 tone pattern. These are estimated by combining the partial effects of the factor smooth for \texttt{word} with the general contour of the T2–T3 tone pattern produced by female speakers (cf.\ Figure~\ref{fig:genderXtone}, panel~1). 
    The red dashed line is reproduced from Figure~\ref{fig:genderXtone}, panel~1 for reference. 
    The grey dashed vertical line indicates the normalized syllable boundary, averaged across tokens for each word type, and the grey dashed horizontal line marks the reference at $y = 0$. 
    The ``n=" label in the upper-right corner of each panel shows the token count for each word type. Panels are ordered by token count, from highest to lowest.}}
    \label{fig:word23}
\end{figure}

\begin{figure}[t]
    \centering
    \includegraphics[width=\textwidth]{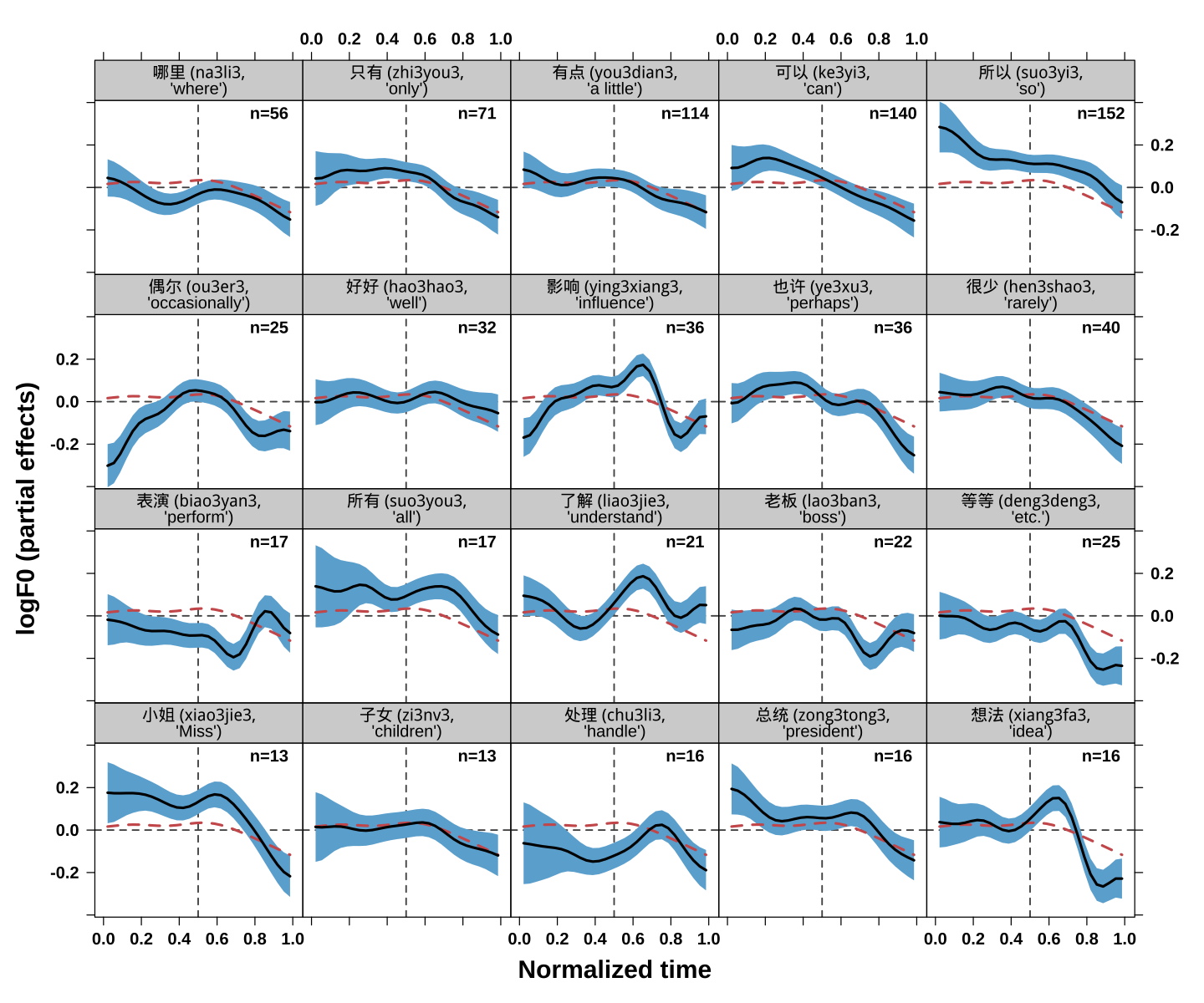}
    \caption{{\color{black}Predicted pitch contours for words with the T3–T3 tone pattern, estimated by combining the partial effects of the factor smooth for \texttt{word} and the general contour of T3-T3 tone pattern produced by female speakers (cf.\ Figure~\ref{fig:genderXtone}, panel~3, shown in orange for reference).}}
     \label{fig:word33}
\end{figure}

\noindent
{\color{black}
Within the very same tone pattern, many words have their own F0 fingerprints. Figure~\ref{fig:word23} and ~\ref{fig:word33} present the partial effect of \texttt{word} for all word types in our dataset, for T2-T3 and T3-T3 respectively. The first panel displays words with lexical rising T2-T3 tone pattern, and the second panel shows words with sandhi rising T3-T3 tone pattern. The red dashed contour presents the general contour of female speakers (c.f panel 1 and panel 3 in Figure~\ref{fig:genderXtone}), provided here for reference. Some pitch contours align closely with the expected canonical contour, see, e.g., 而且\ (\textit{er2qi3}), 如果\ (\textit{ru2guo3}, `if') and 国小\ (\textit{guo2xiao3}, `elementary school') in the first panel, and 偶尔\
(\textit{ou3er3}, `occasionally'), 影响\ (\textit{yingxiang3}, `influence') and 想法\ (\textit{xiang3fa3}, `idea') in the second panel. However, many words show rather different patterns, such as a falling contour (e.g., 朋友\ (\textit{peng2you3}, `friends'), 为止\ (\textit{wei2zhi3}, `until') and 台语\ (\textit{tai2yu3}, `Taiwanese') in the first panel, and 可以\ (\textit{ke3yi3}, `can'), 所以\ (\textit{suo3yi3}, `so'), and 很少\ (\textit{hen3shao3}, `rarely') in the second panel).  A level contour is observed for 学长\ (\textit{xue2zhang3}, `senior') in the first panel and 好好\ (\textit{hao3hao3}, `well') in the second panel.}
{\color{black}
It is clear that, at level of individual words, differences in pitch contours are present for both the first and the second syllable. Although, on average, tone sandhi results in a high and level tone for the first syllable and a falling tone for the second syllable, the way in which F0 is actually realized differs from word to word, and involves both syllable within words. 
}\footnote{
{\color{black}
The studies by \textcite{Chuang:Bell:Tseng:Baayen:2025,jin2025new,lu2026realization}, provide strong evidence that these kinds of word-specific tonal contours reflect semantics. First, when word is replaced by sense in GAMs, prediction accuracy improves. Second, pitch contours can be predicted from contextualized embeddings with accuracies that are substantially higher than permutation baselines. This is strong evidence that word-specific pitch signatures originate at least in part from lexis and grammar.
}
}

\subsection{Variable Importance}

\begin{figure}[htbp]
    \centering
    \begin{subfigure}[b]{0.54\textwidth}
        \centering
        \includegraphics[width=\linewidth,height=7.5cm]{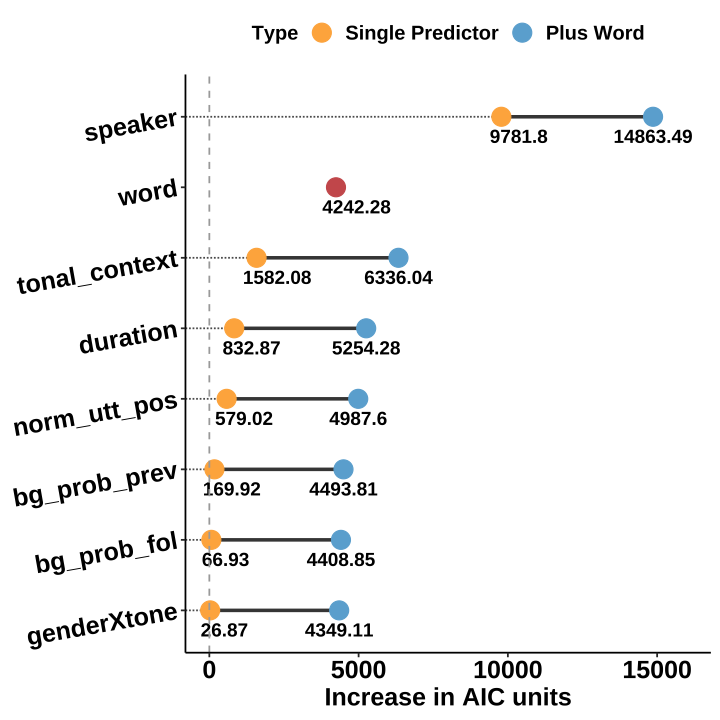}
    \end{subfigure}
    \hfill
    \begin{subfigure}[b]{0.45\textwidth}
        \centering
        \includegraphics[width=\linewidth, height=7cm]{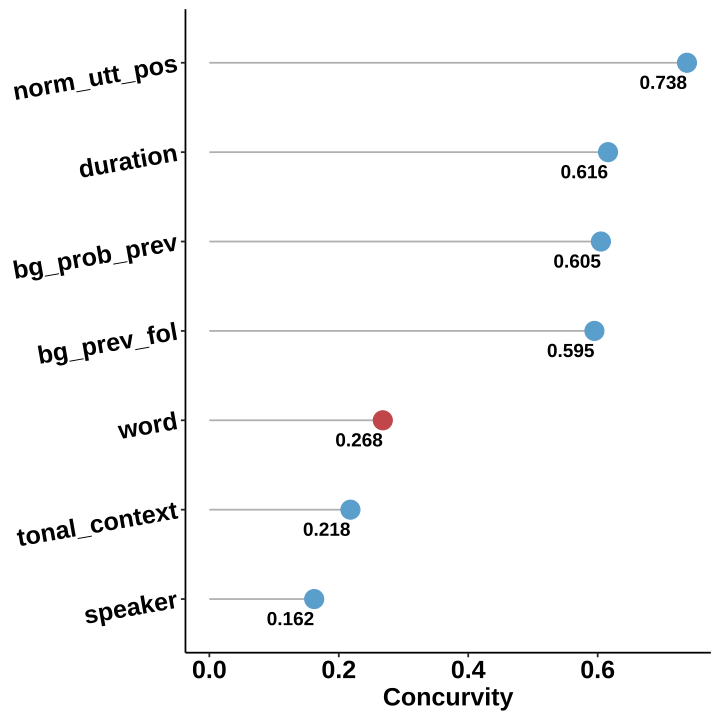}
    \end{subfigure}
    \caption{Variable importance (left) and concurvity (right) for selected terms in the GAMM fitted to the pitch contours of T2-T3 and T3-T3 words.  Variable importance is evaluated with the increase in AIC when a predictor is withheld from the model specification (orange dots).  In the left panel, the blue dots show the additional increase in AIC when a predictor is withheld together with word. Concurvity scores are calculated for the GAMM with all predictors included (summarized in Table~\ref{tab:gam.summary}). {\color{black}The predictor \texttt{word} is highlight in red.}}
    \label{importance}
\end{figure}

\noindent
In order to assess the variable importance of the predictors, we calculated the increase in AIC when a predictor is withheld from the model specification. The left panel of Figure~\ref{importance} clarifies that \textit{speaker} and \textit{word} have the greatest variable importance.  The blue dots in this panel clarify that when a predictor and \texttt{word} are withheld jointly, the additional increase in AIC is roughly the same for all predictors, indicating that the effect of \texttt{word} cannot be traced back to the effects of other predictors.    

The right panel of Figure~\ref{importance} presents the concurvity scores for the predictors in the left panel.  These scores, which range between 0 and 1, indicate to what extent the effect of a predictor is confounded with the effect of other predictors in the model.  Concurvity scores are lowest for speaker, tonal context and word, which allows us to conclude that the effects of these predictors are mostly independent of the effects of the other predictors in the model.  {\color{black}Conversely, concurvity scores are much higher for duration, word position, and bigram probabilities. This is perhaps unsurprising given that, for instance, duration tends to increase as an utterance unfolds ($r = 0.23, p < 0.0001$),} whereas bigram probability conditioned on the following word decreases with a word's position in the sentence ($r = -0.29, p < 0.0001$).

\section{\MakeUppercase{The effect of word}}

\noindent
We have seen that \texttt{word} is an important predictor of how pitch contours are realized. This raises the question of what this effect actually reflects. In what follows, we first consider the possibility that it is actually a lexical frequency effect, given that \textcite{yuan_3rd_2014} reported stronger tone sandhi effects for lower-frequency T3-T3 words. We then consider {\color{black}whether the word effect arises from the segmental make-up of words.} Finally, semantics co-determines tonal realizations, as suggested by findings reported in \textcite{Chuang:Bell:Tseng:Baayen:2025, lu2026realization}.

\subsection{Word or Word Frequency?}

\noindent
To clarify whether the effect of word can be traced back to an effect of frequency of use, we considered two word frequency measures: \texttt{spoken\_freq} (the frequency of word in the current corpus of spoken Taiwan Mandarin) and \texttt{written\_freq} (the frequency of word from the written Sinica corpus).    

{\color{black}
\begin{description}
    \item[\texttt{spoken\_freq}] Number of occurrences of a word type in the spoken Taiwan corpus used in the present study. 
    \item[\texttt{written\_freq}] Number of occurrences of a word type in the Sinica Corpus of Taiwan Mandarin, a large-scale text containing over 10 million words \parencite{ma__2001}. 
\end{description}
}

As shown in Table~\ref{tab:freq}, including spoken frequency (i.e., an interaction of frequency by normalized time) improves model fit by {\color{black}853.35} AIC units and written frequency (with the same interaction) improves model fit by {\color{black}961.06} AIC units. Nevertheless, frequency is confounded with word type. For example, there are 39 distinct written frequency values for 40 word types. This raises the question of whether the effect of word is actually an effect of frequency. Two observations argue against this possibility. First, adding \texttt{word} to the model that already includes \texttt{written\_freq} or \texttt{spoken\_freq} leads to a further substantial improvement in model fit {\color{black}by 4,259.56 and 4,262.72 AIC units respectively}. {\color{black}However, when frequency and word are both included, the model summary does not support the smooth term for \texttt{spoken\_freq} ($p=0.1324$) and its interaction with normalized time ($p=0.5676$). The same holds for the interaction between normalized time and \texttt{written\_freq} ($p=0.5257$). Notably, the concurvity of both \texttt{spoken\_freq} and \texttt{written\_freq} is 1, indicating very high collinearity.} Second, a simpler model that includes word, but neither \texttt{written\_freq} nor \texttt{spoken\_freq}, provides a fit that is almost as good as a model with word and frequency-related predictors (see Table~\ref{tab:freq}, Row 4-6). 

{\color{black}In other words, although models with frequency measures tend to have slightly lower AIC, the partial effects of frequency are not well-supported. The evidence for frequency as a predictor is therefore ambiguous, but including word as a predictor clearly has a stronger effect on tonal realizations.} This allows us to conclude that word has an independent effect on tonal realizations that cannot be reduced to lexical frequency. 


\begin{table}[htbp]
\centering 
\tablesize
\caption{AIC scores of GAMM models fitted with \texttt{word}, \texttt{spoken\_freq} (frequency in the Taiwan spoken corpus), or \texttt{written\_freq} (frequency in the larger written Sinica Corpus of Taiwan Mandarin), either replacing \texttt{word} or added as an additional predictor. {\color{black}The ``best-fit GAMM" refers to the model we fit with the model specification presented in Section~\ref{sec:results}. All models are fitted using the word dataset (see Table~\ref{tab:data}).}}
\begin{adjustbox}{width=\textwidth}
\begin{tabular}{lrr}
\toprule
\textbf{Model} & \textbf{AIC} & \textbf{$\Delta$AIC} \\ 
\midrule
\textbf{Baseline model: all other predictors} (word dataset) & -274052.08 & -- \\ 
\texttt{+ spoken\_freq}  & -274905.43 & -853.35 \\ 
\texttt{+ written\_freq} & -275013.14 & -961.06 \\ 
\texttt{+ word} (best-fit GAMM) &  \textbf{-278294.36} & \textbf{-4242.28} \\ 
\texttt{+ written\_freq + word} &  -278311.64 & -4259.56 \\ 
\texttt{+ spoken\_freq + word} & -278314.80 & -4262.72 \\ 
\bottomrule
\end{tabular}
\end{adjustbox}
\label{tab:freq}
\end{table}

{\color{black}
\subsection{Word or Segments?}

\noindent
In addition, \textcite{Chuang:Bell:Tseng:Baayen:2025} reported that word type is a strong predictor of tonal realization, outperforming a range of segmental properties of words' forms such as vowel height, consonant onset, and rhyme structure. In the present study, we therefore treat \texttt{word} as a critical predictor {\color{black}in the best-fit GAMM}, avoiding issues of collinearity that arise when including segmental properties as predictors. Nonetheless, we present a series of model comparisons to assess the relative effect of segments. Following \textcite{Chuang:Bell:Tseng:Baayen:2025}, we include the following segment-related variables:

\begin{description}
    \item[vowel height] A categorical predictor encoding the height of the vowel for the first and second syllables separately: \textit{vowel1\_height} and \textit{vowel2\_height}. Each is a factor with five levels: \textit{high}, \textit{mid}, \textit{low}, \textit{low-high}, and \textit{mid-high}.
    
    \item[onset type] The phonological category of the initial consonant of each syllable. We include \textit{onset1\_type} and \textit{onset2\_type}, each coded as a factor with seven levels: \textit{aspirated-affricate}, \textit{aspirated-stop}, \textit{unaspirated-affricate}, \textit{unaspirated-stop}, \textit{voiceless-fricative}, \textit{voiced}, and \textit{null}.
    
    \item[rhyme] The internal structure of the syllable rime, based on the presence of a prenuclear glide and/or a nasal coda. We distinguish four rhyme types: \textit{V} (vowel only), \textit{GV} (glide + vowel), \textit{VN} (vowel + nasal), and \textit{GVN} (glide + vowel + nasal). These are encoded separately for each syllable as \textit{rhyme1} and \textit{rhyme2}.
\end{description}

We began with a baseline model that included all predictors except \texttt{word}, specifically gender, tone pattern, duration, tonal context, word position in the utterance, bigram probability, and speaker. To examine the contribution of segment-related predictors, we incrementally added them to the baseline model. As shown in Table~\ref{tab:segments}, each addition gradually reduced the AIC, indicating improved model fit. However, even after accounting for all segment-related variables, the model including \texttt{word} achieved a substantially lower AIC (-278,294.36) than the model combining all segment-related predictors (-277,801.37). 
This finding suggests that by the time all segment-related predictors are included, one has come very close to defining the word. Nonetheless, in line with \textcite{Chuang:Bell:Tseng:Baayen:2025}, the fact that the \texttt{word} predictor still provides a substantial improvement in model fit indicates that it captures additional properties not fully accounted for by the segmental makeup alone.

\begin{table}[htbp]
\centering
\tablesize
\caption{{\color{black}Improvement in model fit, measured by AIC reduction, as segment-related predictors are added to the baseline model. The ``best-fit GAMM" refers to the model we fit with the model specification presented in Section~\ref{sec:results}. All models are fitted using the word dataset (see Table~\ref{tab:data}).}}

\begin{tabular}{lrr}
\toprule
\textbf{Model} & \textbf{AIC} & $\boldsymbol{\Delta}$\textbf{AIC} \\
\midrule
\textbf{Baseline model: all other predictors} (word dataset) & -274052.08& -- \\
+ Onset consonant types (\texttt{onset1\_type}, \texttt{onset2\_type}) & -276183.68& -2131.60 \\
\hspace{1em}+ Vowel height (\texttt{vowel1\_height}, \texttt{vowel2\_height}) & -277107.42& -3055.34\\
\hspace{2em}+ Rhyme structure (\texttt{rhyme1}, \texttt{rhyme2}) & -277801.37 & -3749.29\\
+ Word (\texttt{word}) (best-fit GAMM) & \textbf{-278294.36}& \textbf{-4242.28}\\
\bottomrule
\end{tabular}
\label{tab:segments}
\end{table}
}

\subsection{Word or Word Sense?}

\noindent
\textcite{Chuang:Bell:Tseng:Baayen:2025} reported that replacing word by word sense improved GAM models fitted to the F0 contours of two-character words with T2-T4 tone pattern. We therefore investigated whether word sense is also a superior predictor for the present dataset of words with the T3-T3 and T2-T3 tone patterns. More specifically, we used the same model specification as presented in Section~\ref{sec:results}, but fitted two GAMMs to the sense dataset with 1144 tokens of 46 sense types (see Table~\ref{tab:data}). The first GAMM incorporated by-word factor smooths. In the second GAMM, these by-word factor smooths were replaced with by-sense factor smooths.

As shown in Table~\ref{tab:aic_sense}, replacing the by-word factor smooth with a by-sense factor smooth leads to a substantial further improvement in model fit {\color{black}($\Delta$AIC: -4,472.60 and $\Delta$AIC: -5,549.62)}. Furthermore, the concurvity score for the by-sense smooths ({\color{black}0.296}) was slightly smaller than the corresponding score for the by-word smooths ({\color{black}0.334}), which suggests that replacing word by sense did not lead to a less interpretable model. This result supports the hypothesis that tonal variation is more strongly tied to word sense than to word.

\begin{table}[htbp]
\centering
\tablesize
\caption{{\color{black}Improvement in model fit, measured by AIC reduction, when adding \texttt{word} or \texttt{sense\_type} as predictors to the baseline model. All models are fitted using the sense dataset (see Table~\ref{tab:data}).}}
\begin{tabular}{lrr}
\toprule
\textbf{Model} & \textbf{AIC} & \textbf{$\Delta$AIC} \\ 
\midrule
\textbf{Baseline model: all other predictors} (sense dataset) & -115835.21 & - \\
\texttt{+ word}  & -120307.81 & -4472.60 \\ 
\texttt{+ sense type} & \textbf{-121384.83} & \textbf{-5549.62} \\ 
\bottomrule
\end{tabular}
\label{tab:aic_sense}
\end{table}

\begin{figure}[htbp]
    \centering
    \includegraphics[width=\linewidth]{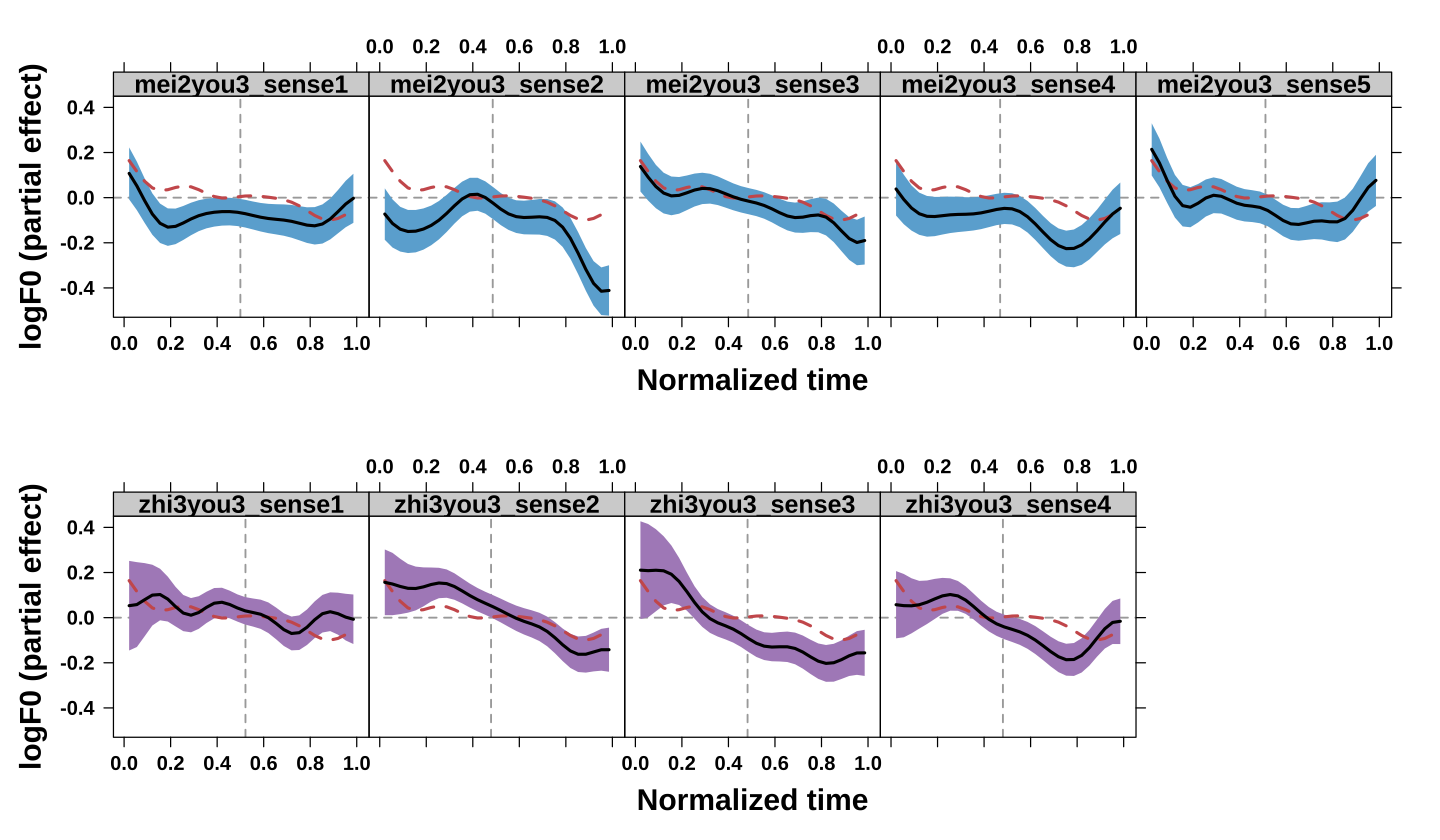}
    \caption{\color{black}{The predicted pitch contour for sense types of the words 没有\ (\textit{mei2you3}, `not have'; upper panel) and  只有\ (\textit{zhi3you3}, `only'; lower panel), based on the \texttt{baseline + sense type} model (see Table~\ref{tab:aic_sense}). 
    The predicted contours are calculated from the partial effect of the factor smooth for sense type, combined with the general contour of T2-T3 or T3-T3 by female speakers (shown as red dashed line). 
    In the upper panel, the sense types (sense 1--5) are: `there is not', `not have', `do not', `be not as...as...', `expressing questioning'. In the lower panel, the sense types (sense 1-- 4) are: `only', `exceptional', `only if', and `just'. For more details on sense types, see Table~\ref{tab:sense_def} in Appendix. 
    Vertical dashed lines indicate the syllable boundaries on the normalized time scale averaged for each sense type, and horizontal dashed lines indicate the $y=0$ reference line.}}
    \label{fig:sense}
\end{figure}

Figure~\ref{fig:sense} presents the predicted pitch contours of different sense types for two word types: 没有\ (\textit{mei2you3}, `none', upper panel) with T2-T3 tone pattern and 只有\ (\textit{zhi3you3}, `only', lower panel) with T3-T3 tone pattern. Although the sense types for 没有\ 
are fairly similar, they show up with fairly distinct tonal contours: a rising contour for sense 1, a dipping contour for sense 5, and a rise-fall for sense 2. The pitch contours for 只有\ 
are more similar, with the strong fall for sense 3 standing out most.
The enhanced precision offered by sense as compared to word provides further evidence for the hypothesis offered by \textcite{Chuang:Bell:Tseng:Baayen:2025} that the fine details of Taiwan Mandarin pitch contours are functional in that they facilitate understanding what words mean given the phonetic signal. For other studies reporting a close relation between semantics and phonetic realization, see \parencite{chuang_processing_2021, gahl_time_2024, lu2026realization,schmitz2022production,Saito2024}.

The observation that words with the same tonal pattern show different tonal realization depending on their semantics dovetails well with usage-based approaches to speech production \parencite{bybee_phonology_2003, hawkins_roles_2003, pierrehumbert_word-specific_2002} as well as examplar-based approaches --- see, for instance,  \textcite{li_complete_2016}, who reported that neutralization of Fuzhou tone sandhi requires a hybrid phonological competence model encompassing both abstract categories and stored exemplars. For evidence based on computational modeling that meaning and pitch are intimately related and learnable, see \textcite{Chuang:Bell:Tseng:Baayen:2025} and \textcite{lu2026realization}.

{\color{black}
\subsection{Cross-validation}

\noindent
If the effect of \texttt{word} is robust, and the model is not overfitting, a GAMM that includes \texttt{word} as a predictor should be more accurate in predicting pitch contours for held out data compared to GAMMs that only has access to control predictors or frequency. To evaluate this, we conducted a cross-validation analysis using the word dataset. The dataset was split into training (90\%) and testing (10\%) subsets, with each word type being represented in both. The following model specifications were considered:

\begin{description}
    \item[Model 0:] \texttt{genderXtone} + \texttt{speaker} (baseline)
    \item[Model A:] \texttt{genderXtone} + \texttt{speaker} + \texttt{word}
    \item[Model B:] \texttt{genderXtone} + \texttt{speaker} + \texttt{duration} + \texttt{word}
    \item[Model C:] All control factors + written frequency (see last row in Table~\ref{tab:freq})
    \item[Model D:] All control factors + all segment-related predictors (see Row~4 in Table~\ref{tab:segments})
    \item[Model E:] All control factors + \texttt{word} (see Section~\ref{sec:results}, the best-fit model)
\end{description}

Model 0 is the simplest and serves as the baseline, with only a by-\texttt{genderXtone} smooth of time and a by-\texttt{speaker} factor smooth. Model A extends the baseline model by including a factor smooth for \texttt{word}. Model B is slightly more complex than model A, with the addition of a smooth of \texttt{duration}. Model C represents the best-performing model among frequency-based models, and Model D is the best-performing model among segments-based models. Model E is the best-fit model we reported, with all control variables together with \texttt{word}.  We expect all models to outperform the baseline model. Furthermore, as the model including segmental predictors (model~D) identifies words in a more complex way than \texttt{word} itself, we expect its accuracy to be similar to that of model E. Here, we note that model~E has lower AIC than model~D, due to model~E being the simpler model (see Table~\ref{tab:segments}). 

Prediction accuracy was assessed by generating predicted F0 contours for the held-out data and calculating the sum of squared errors (SSE)\footnote{The sum of squared errors (SSE) is the sum of the squared differences between observed and predicted values. Smaller values indicate greater predictive accuracy.}. Each model was evaluated across 50 random training–testing splits. Figure~\ref{fig:cross} (a) displays boxplots of the SSE differences between Model~0 (baseline) and each comparison model. Positive values indicate a lower error than the baseline, with larger positive values reflecting greater improvements in prediction accuracy.  The t-tests indicated that for each model, the differences are highly unlikely to be zero (all $p < 0.0001$).

\begin{figure}[t]
    \centering
    \begin{subfigure}[b]{0.33\textwidth}
        \centering
        \includegraphics[width=\linewidth]{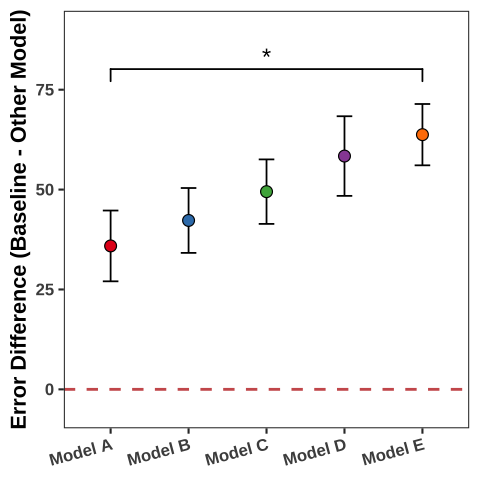}
        \caption{}
    \end{subfigure}
    \hfill
    \begin{subfigure}[b]{0.66\textwidth}
        \centering
        \includegraphics[width=\linewidth]{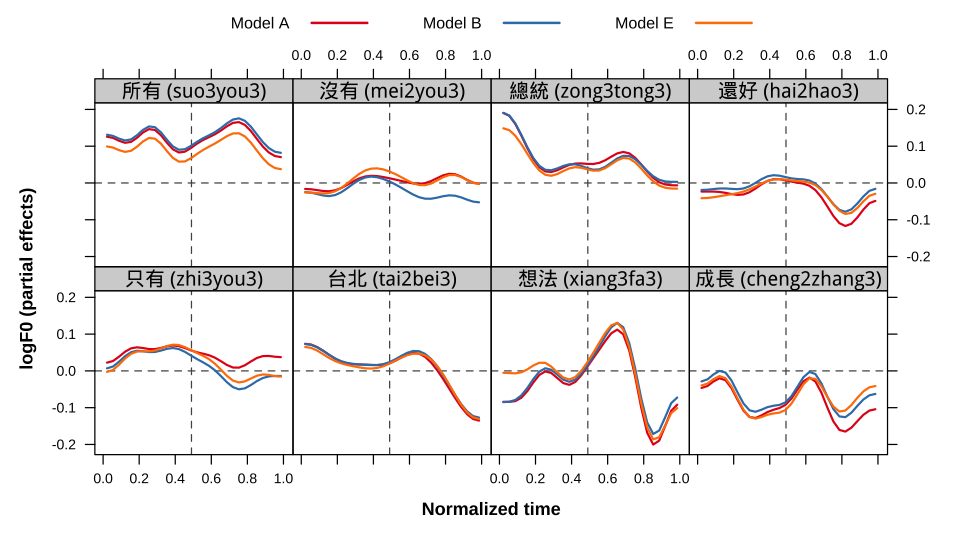}
        \caption{}
    \end{subfigure}
    \vskip 0.25em
    \begin{subfigure}[b]{1\linewidth}
        \centering
        \includegraphics[width=\linewidth]{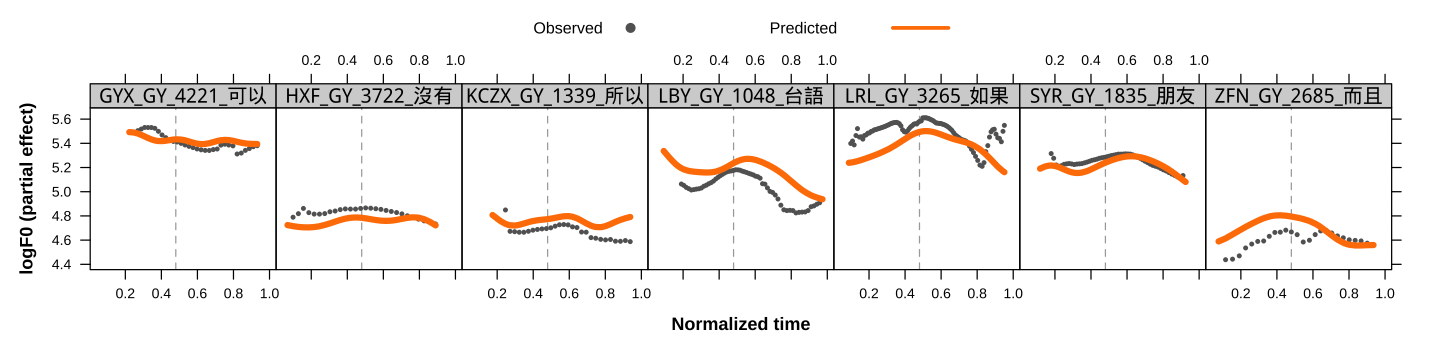}
        \caption{}
    \end{subfigure}
    
    \caption{
    Panel (a): Predictive accuracy of five models over 50 cross-validation runs for the word dataset, each compared with the baseline model. Boxplots show the SSE differences between Model~0 (baseline) and other models. Positive values indicate lower errors than the baseline, with larger distances from zero reflecting greater improvements in prediction accuracy.
    Panel (b) illustrates the partial effect of the \texttt{word} factor smooth for Models~A, B, and E. 
    Panel (c) shows predicted pitch contours for high-frequency tokens previously excluded, as predicted by Model~E. Grey dots represent observed data, and the red line shows the predicted contour.}
    \label{fig:cross}
\end{figure}

First, consider Models~A, B, and E, which all include \texttt{word} but vary in model complexity.  If the observed word-specific contours were an artifact of overfitting, simpler models would not show these effects. As shown in Figure~\ref{fig:cross} panel (a), Model~E outperforms Model A, indicating that including additional control variables improves predictive accuracy ($p = 0.0195$). 

Second, consider Models~C, D and E. While Model E has a slightly higher mean SSE difference than Model~D (Model~D: 58.39; Model E: 63.74), t-test suggests that this difference is not statistically significant ($p = 0.6713$). As shown in Table~\ref{tab:segments}, the AIC of the full model with word (-278,294.36) is much lower than that of Model~D (-277,801.37). The cross-validation results clarify that the model with word is simpler than the segmental model without a loss of prediction accuracy. We also note that Model~E has slightly higher SSE advantage than model~C (model~C: 49.48).

Panel (b) of Figure~\ref{fig:cross} presents for a selection of words the pitch contours predicted by models~A, B, and E, using the full dataset. The predicted word-specific pitch contours are very similar across models. In other words, all models provide reliable predictions as long as word is included, with, as shown in panel (a), more complex models offering some improvements in prediction.

Recall that in Section~\ref{sec:data}, we randomly sampled 200 tokens for high-frequency words, in order to avoid model prediction being biased towards the high-frequency words.  In order to clarify how well the model generalizes to the tokens that we excluded, we took the best word model (model E), trained it on the full dataset, and examined prediction accuracy for 200 novel tokens (previously excluded) for the 7 most frequent words.  Figure~\ref{fig:cross} panel (c) presents observed pitch values and predicted pitch contours for one randomly selected token of each of these seven frequent word types. Predicted contours are fairly close to the observed data points, providing further confidence in the quality of the full model.
}

{\color{black}
\subsection{Word and Neutralization} 

\noindent
Having presented the full GAM model for our data, and having shown this model is not overfitting, we examine once more whether there is any evidence for incomplete neutralization of the T3-T3 pattern in the sense dataset. As a first step, we compared three GAMs, one with separate contours for each combination of speaker gender and tone pattern, one with separate smooths for the two tone patterns but without an interaction with gender, and one with only a general smooth for time.  The AIC scores for these three models were {\color{black}very close (-121,384.8, -121,349.7 and -121,343.1)}, so we prefer the simplest model with only a general smooth for time that is the same across genders and tone patterns.

As a second step,  we considered the possibility that incomplete neutralization is visible in our dataset when word or word sense are not included as predictors. {\color{black}In the word dataset, when the factor smooth for word is removed from the model specification, a general smooth by \texttt{genderXtone} is supported by a lower AIC (-274,070.4) compared to a model that only includes a general smooth for normalized time (-273,947.0, $\Delta$AIC: -123.42). In the sense dataset, when the factor smooth for sense type is removed from the model specification, we observe the same: a general smooth by \texttt{genderXtone} is supported by a lower AIC (-115,835.2) compared to a model that only includes a general smooth for normalized time (-115,742.9, $\Delta$AIC: -92.35)}. However, all models without word or word sense as predictors are substantially inferior to the models that include these predictors (AIC differences increase by two orders of magnitude). We conclude that there is no evidence in our data for incomplete neutralization in conversational Taiwan Mandarin, once word or word sense are in place as proper controls.}

\section{\MakeUppercase{General discussion}}

\noindent
Previous literature has reported that there is an incomplete neutralization process of Tone 3 sandhi in Beijing Mandarin \parencite{xu_contextual_1997}. \textcite{xu_contextual_1997} showed for laboratory speech that T2-T3 and T3-T3 tones exhibit very similar contours, characterized by an  initial slight fall, followed by a clear rise, and then a large fall.  At the same time, the contour for T2-T3 was found to have a slightly higher pitch across most of the two syllables.  \textcite{yuan_3rd_2014} reported for standard Mandarin connected speech (telephone conversations and news broadcasts) that the sandhi T3 in T3-T3 words differs with the T2 in T2-T3 in terms of the magnitude and time span of F0 rising. However, the difference between T2-T3 and T3-T3 was reported not to be statistically significant in Taiwan Mandarin laboratory speech \parencite{peng_lexical_2000}.


We investigated the realization of Tone 3 sandhi in conversational speech recorded in Taiwan. Above (in Figure~\ref{fig:example}), we presented  a sample of pitch contours for bi-character words, and observed not only that the description given by \textcite{xu_contextual_1993} holds only for some tokens, but also that words can have very different actual pitch contours.  The difference between T2-T3 and T3-T3 reported by \textcite{yuan_3rd_2014} is also not clearly visible in Figure~\ref{fig:example}. For example, the logarithm of the ratio of between the F0 at syllable offset and minimum F0 in their data of telephone conversation appears to have always been positive (see their Figures~1), which means that the F0 at the syllable offset is always greater than the minimum F0. But in our data on Taiwan Mandarin, there are tokens for which the logarithm of  this ratio would be 0 (see the token  LJS\_GY\_6870\_媒体\ in Figure~\ref{fig:example} which has a fall in its first syllable).  

We made use of the generalized additive model \parencite{wood_generalized_2017} to predict pitch as a function of (normalized) time. Unlike previous studies, we have taken into account a wide range of factors known to co-determine the realization of pitch (gender, {\color{black}duration}, word position in the utterance, conditional bigram probabilities, tonal context, speaker, and word). A further, new control variable that we took into account was word sense \parencite{Chuang:Bell:Tseng:Baayen:2025}. With these factors controlled for, for spontaneous conversational Taiwan Mandarin, the difference between T2-T3 and T3-T3 in Taiwan Mandarin is not statistically significant, {\color{black}for female speakers. For male speakers, T2-T3 might be higher than T3-T3 in the beginning of the syllable, but the statistical evidence is ambiguous.
Therefore, in line with previous research, our findings provide strong support for the  neutralization of Tone 3 sandhi in informal Taiwan Mandarin being complete, certainly for female speakers, and probably also for male speakers.}

This disparity with the study by \textcite{yuan_3rd_2014} can be attributed to several factors, such as differences in the register of the speech (telephone speech and news broadcasts as opposed to spontaneous face-to-face conversation), differences between the dialects of Mandarin sampled (Beijing Mandarin as opposed to Taiwan Mandarin, which is influenced by Southern Min), methodology (point measurements as opposed to modeling full F0 contours), and the absence of word or word sense as controls.  It is of course possible that in Beijing Mandarin, the tone sandhi is indeed incomplete even when word or word sense are included as predictors.  We leave this issue for future research.

{\color{black}In our GAMM analysis, once word is controlled for, there is no clear evidence of incomplete neutralization. First, one might argue that once frequency of occurrence is taken into account, evidence for incomplete neutralization would emerge more clearly. However, word outperforms lexical frequency as a predictor of the F0 contour. 
Second, by the time all segment-related predictors such as onset consonant, vowel height, and rhyme structure are included, the model nearly captures the identity of the word itself. Yet, the effect of word cannot be fully reduced to the combined influence of these segment-related variables.
Furthermore, when word-specific F0 components are replaced by sense-specific F0 components in a further analysis, model fit improves further. Clearly, word is an intrinsic predictor of pitch contours, and with this essential predictor in place, for Taiwan conversational Mandarin as recorded in Taiwan, the evidence for incomplete neutralization is extremely weak.}

In summary, our analyses indicate, first, that in conversational Taiwan Mandarin, Tone 3 preceding Tone 3 completely assimilates to Tone 2, and second, that F0 contours vary systematically with word sense. Apparently, disyllabic words have their own F0 fingerprint in Taiwan Mandarin Chinese conversational speech, irrespective of whether the tone of the first character is T2 or T3.  

{\color{black}Further investigations of tonal realizations in Taiwan Mandarin, focusing on 20 tone patterns \parencite{lu2026realization} and monosyllables \parencite{jin2025new}, replicate the word-specific tonal realizations originally observed for the T2-T4 pattern by \textcite{Chuang:Bell:Tseng:Baayen:2025}. Moreover, these studies show that pitch contours can be predicted from meaning by computational modeling with above-chance accuracy.}

The current study extends research on Tone 3 sandhi from careful speech to spontaneous conversations. Following \textcite{Chuang:Bell:Tseng:Baayen:2025}, we have harnessed the power of the generalized additive model, which made it possible to model complete F0 trajectories as a function of (normalized) time, instead of point estimates gauging aspects of F0 contours (such as the magnitude, and the duration, of the F0 rise). {\color{black}We note that additional factors could be controlled for when modeling F0 contours, but these would primarily serve to fine-tuning the model (see the supplementary material). For example, bigram probabilities exerted relatively smaller effects compared to other predictors such as word and speaker. Importantly, the effects of word and sense remain robust in the GAMM analysis.}

{\color{black}
One direction for future research is to investigate the realization of T2-T3 and T3-T3 tone patterns in spontaneous conversations in other dialects of Mandarin Chinese, such as Beijing Mandarin.} A topic for further investigation is the observed word-specific or sense-specific pitch signatures that the GAMM reports for Taiwan Mandarin. Are these pitch signatures completely dialect specific, or do they generalize to other dialects? {\color{black}Preliminary results of ongoing research \parencite{lu2025variations} suggest that, as expected for different dialects, similarities gang up with dissimilarities.}
{\color{black}Another further extension would be to examine pitch signatures in the higher registers of spoken Mandarin, such as found in broadcast news speech, and audio books.}

\newpage
\input{appendix_revised}

\clearpage

\vspace{2\baselineskip}
\input{appendixR2}
\clearpage

\vspace{2\baselineskip}
{\centering \MakeUppercase{Notes}\par}
\vspace{-3\baselineskip}
\theendnotes

\vspace{2\baselineskip}  
\normalsize
\printbibliography

\newpage
\myctitle{形式和意义共同决定声调在台湾普通话自发性语流中的实现：以T2-T3和T3-T3的三声连续变调为例}

\mycauthor{卢语欣}{图宾根大学}{}
\mycauthor{庄育颖}{国立台湾师范大学}{}
\mycauthor{R. Harald Baayen}{图宾根大学}{}

\chineseabstract
{\noindent
在标准汉语中，当两个三声（上声）连续出现时，第一个三声变为第二声（升调），这称为上声连读变调。过去研究指出，上声连读变调可能是一种不完全同化，即上声连读变调的升调仍然与阳平的升调存在一定区别。尽管普通话的上声连读变调现象在系统性控制的语音实验资料（Xu，1997）和正式语域的北京普通话（Yuan \& Chen，2014）中被广泛研究，但其在自发性语流中的实现以及语境因素对声调的影响尚不清楚。
本研究探讨了台湾普通话的自发性语流中具有T2-T3和T3-T3声调模式的双字词。我们使用广义相加混合模型（GAMM，Wood，2017）来检验声调轮廓随标准化时间的变化。在模型中，我们考虑了多种影响声调轮廓的因素，包括性别、时长、词语位置、双字组概率、相邻声调、语者，以及新的变项：词与词意（Chuang et al.，2025）。分析显示，在台湾普通话自发性语流中，当加入词（或词意）作为变项后，T3-T3与T2-T3的区别便不显著，亦即上声连读变调是一种完全同化。}

\chinesekeywords{
\noindent
\textbf{三}声连续变调\ \space \space \textbf{声}调同化\ \space \space \textbf{广}义线性相加模型\ \space \space \textbf{词}汇特定调音实现\ 
}

\end{document}

%% file: appendix_revised.tex
{\centering
\MakeUppercase{Appendix} \par
}
\vspace{1\baselineskip}

\renewcommand{\thetable}{A.\arabic{table}}
\setcounter{table}{0} 

\renewcommand{\thefigure}{A.\arabic{figure}}
\setcounter{figure}{0}

\begin{table}[H]
\caption{Examples of tokens and the utterances in which they occur, grouped by word type.}
\centering
\tablesize
\begin{tabular}{p{5cm} p{9cm}}
\toprule
\textbf{Token} & \textbf{Utterances} \\
\midrule
\multicolumn{2}{l}{\textbf{媒体\ (\textit{mei2ti3}, `media')}} \\
\addlinespace[0.3em]
CHT\_GY\_1008\_媒体 & 台湾\_媒体\_就\_是\_这样\_啊\ (Taiwan media is just like this) \\
LJS\_GY\_6870\_媒体 & 台湾\_媒体\ (Taiwan media) \\
LJS\_GY\_8814\_媒体 & 然后\_媒体\_要\ (Then the media wants to \ldots) \\
LJS\_GY\_8934\_媒体 & 媒体\_最\ (Media is the most \ldots) \\

\addlinespace[1em]
\multicolumn{2}{l}{\textbf{了解\ (\textit{liao3jie3}, `to know')}} \\
\addlinespace[0.3em]
ZWH\_GY\_1763\_了解 & 其实\_我\_太\_了解\_那\_个\_体育\ (Actually, I don't quite understand that sport) \\
GYX\_GY\_4564\_了解 & 不\_太\_了解\_啦\_因为\ (I don't really understand, because...) \\
KCZX\_GY\_1537\_了解 & 我\_依\_我\_了解\_是\_他们\ (According to my understanding, it's them) \\
LCX\_GY\_4771\_了解 & 比较\_能够\_了解\ (Better able to understand) \\

\addlinespace[1em]
\multicolumn{2}{l}{\textbf{子女\ (\textit{zi3nü3}, `children')}} \\
\addlinespace[0.3em]
GYX\_GY\_982\_子女 & 然后\_子女\_呢\ (So the children \ldots) \\
LCX\_GY\_4382\_子女 & 以前\_就是\_对\_子女\_比较\_没有\_什么\_教育\_上\_要\ (In the past, there wasn't much educational requirement for children) \\
LJS\_GY\_9\_子女 & 子女\_这\_个\_项目\ (This item about children) \\
LJS\_GY\_749\_子女 & 子女\_这\_一\_这\_这\_一些\_事情\_啊\ (These \ldots things about children) \\
LJS\_GY\_867\_子女 & 子女\_这\_个\_项目\_里面\ (In this item about children) \\
LJS\_GY\_1240\_子女 & 子女\ (Children) \\
LJS\_GY\_1325\_子女 & 子女\_每\_一\_人\ (Each child) \\
LJS\_GY\_1345\_子女 & 子女\ (Children) \\
LJS\_GY\_1355\_子女 & 教育\_嘛\_那\_子女\ (Education, well, then the children...) \\
LJS\_GY\_1556\_子女 & 子女\_会\_反弹\_然后\_你\_还\_一而再\_一而再\_修正\_你\ (Children will push back, and you keep adjusting yourself again and again) \\
LJS\_GY\_1609\_子女 & 当\_父母\_发现\_子女\_有\ (When parents discover that their children have...) \\
WZW\_GY\_2104\_子女 & 子女\ (Children) \\
\bottomrule
\end{tabular}
\label{tab:tokens_utterances}
\end{table}

\begin{table}[htbp]
\centering
\tablesize
\caption{\color{black}{Sense types presented in Figure~\ref{fig:sense}, along with their definitions and example contexts including English translations.}}
\begin{tabular}{p{3cm} p{4cm} p{7cm}}
\toprule
\textbf{Sense Type} & \textbf{Sense Definition} & \textbf{Context} \\
\midrule
\multicolumn{2}{l}{\textbf{没有\ (\textit{mei2you3}, `not have')}} \\
\addlinespace[0.3em]
没有\_sense1 & 后述对象不存在\ (Indicates the nonexistence of the following object.) & 会干部才来找我因为我们已经\textless 没有\textgreater 人了然后才来找剩下来\ (The officials only came to find me because we already \textless no\textgreater people left, then they came to find the remaining ones.) \\

没有\_sense2 & 不拥有后述对象\ (Indicates the lack of possession of the following object.) & 跟同学在一起可是这些同学都是\textless 没有\textgreater 男朋友的喔因为我们都\ (Together with classmates, but these classmates all \textless don’t have\textgreater boyfriends because we all...) \\

没有\_sense3 & 表否定事件的发生\ (Indicates the denial of an event occurrence.) & 经历过但是现在就已经就\textless 没有\textgreater 再这样了已经度过了应该说\ (Experienced it but now it \textless no longer\textgreater happens like that, it has passed, so to speak.) \\

没有\_sense4 & 表未达后述的程度\ (Indicates not reaching the degree described later.) & 就是教他其实他怎么讲功课\textless 没有\textgreater 很好那英文也没有很好\ (Teaching him, actually, his homework \textless isn’t\textgreater very good, and his English is not very good either.) \\

没有\_sense5 & 表疑问的语气\ (Indicates a questioning tone.) & 他好处是咬字模糊不清嘛\textless 没有\textgreater 啦因为每次我喜欢他每\ (His advantage is that his enunciation is unclear, \textless isn’t it\textgreater, because every time I like him every...) \\

\multicolumn{2}{l}{\textbf{只有\ (\textit{zhi3you3}, `only')}} \\
\addlinespace[0.3em]
只有\_sense1 & 表后述对象独有的\ (Indicates something unique to the following object.) & 哪然后说那地震是怎样是\textless 只有\textgreater 你们家的桌子会摇还是地板\ (Then said about the earthquake, was it \textless only\textgreater\ your family’s table that shook or the floor?) \\

只有\_sense2 & 表从同类中指出例外的对象\ (Indicates an exception from a group of similar items.) & 民俗活动没什么印象大概\textless 只有\textgreater 一些庙可是他们应该不会拜\ (No strong impression of folk activities, probably \textless only\textgreater\ some temples, but they probably wouldn’t worship.) \\

只有\_sense3 & 表须具备的充分条件\ (Indicates a necessary and sufficient condition.) & 如果我以后东西真的真的\textless 只有\textgreater 当教师之后能做比如说\ (If in the future, really, really \textless only\textgreater\ after becoming a teacher can I do such things, for example.) \\

只有\_sense4 & 表对后述事物做评估，表示比预期小或少\ (Indicates an evaluation that is smaller or less than expected.) & 个怀疑那我相信的态度其实\textless 只有\textgreater 百分之六十到七十那我会对\ (I suspect that my belief attitude is actually \textless only\textgreater\ 60 to 70 percent, then I will\ldots) \\
\bottomrule
\end{tabular}
\label{tab:sense_def}
\end{table}

%% file: appendixR2.tex
{\centering
\MakeUppercase{Supplementary materials} \par
}

\vspace{1\baselineskip}

To further examine potential gender modulations of the effects reported in the main text, we fitted an extended GAMM that includes by-gender interactions for predictors such as \texttt{duration} and \texttt{bigram probabilities} (both preceding and following). This specification further includes interactions by gender, which enables us to capture how the effect of predictors vary differently for male and female speakers. The model specification for the extended GAMM is as follows:

\begin{tabbing}
mmmmm\=mm\= \kill
\texttt{logF0}  \> $\sim$ \> \texttt{genderXtone +} \\
      \> \> \texttt{s(normalized\_t, by=genderXtone) +} \\
      \> \> \texttt{s(duration, k=5) +} \\
     \> \> \textbf{\texttt{ti(normalized\_t, duration, by=gender) +} } \\
      \> \> \texttt{s(normalized\_utt\_pos, k=6) +} \\
      \> \> \texttt{ti(normalized\_t, normalized\_utt\_pos) +} \\
       \> \> \texttt{s(bg\_prob\_prev, k=5) +} \\
      \> \> \textbf{\texttt{ti(normalized\_t, bg\_prob\_prev, by=gender) +}} \\
       \> \> \texttt{s(bg\_prob\_fol, k=5) +} \\
      \> \> \textbf{\texttt{ti(normalized\_t, bg\_prob\_fol,  by=gender) +}} \\
        \> \> \texttt{s(normalized\_t, tonal\_context, bs=`fs', m=1) +} \\
       \> \> \texttt{s(normalized\_t, speaker, bs=`fs', m=1) +} \\
       \> \> \texttt{s(normalized\_t, word, bs=`fs', m=1)} \\
\end{tabbing}

As shown in Table~\ref{tab:gam}, smooth terms such as \texttt{ti(normalized\_t, \linebreak duration)} and \texttt{ti(normalized\_t, bg\_prob\_prev)} show significant gender-based modulations. The effect of \texttt{bg\_prob\_fol} remains non-significant, but its interaction with time by gender is significant.

\begin{table}[htbp]
\centering
\tablesize
\caption{Model summary of the extended GAMM with further by-gender modulations.} 
\begin{adjustbox}{width=\textwidth}
\begin{tabular}{lrrrr}
   \hline
A. parametric coefficients & Estimate & Std. Error & t-value & p-value \\ 
  (Intercept) & 5.2692 & 0.0261 & 202.2543 & $<$ 0.0001 \\ 
  genderXtonemale.23 & -0.5073 & 0.0319 & -15.8854 & $<$ 0.0001 \\ 
  genderXtonefemale.33 & -0.0232 & 0.0213 & -1.0913 & 0.2751 \\ 
  genderXtonemale.33 & -0.5352 & 0.0383 & -13.9873 & $<$ 0.0001 \\ 
   \hline
B. smooth terms & edf & Ref.df & F-value & p-value \\ 
  s(normalized\_t):genderXtonefemale.23 & 5.0544 & 5.2977 & 7.1619 & $<$ 0.0001 \\ 
  s(normalized\_t):genderXtonemale.23 & 2.0414 & 2.1944 & 6.8479 & 0.0009 \\ 
  s(normalized\_t):genderXtonefemale.33 & 4.3190 & 4.6610 & 4.9280 & 0.0003 \\ 
  s(normalized\_t):genderXtonemale.33 & 2.3984 & 2.6240 & 2.1544 & 0.0896 \\ 
  s(duration) & 3.4623 & 3.8405 & 39.2308 & $<$ 0.0001 \\ 
  ti(normalized\_t,duration):genderfemale & 22.1105 & 23.9876 & 32.0453 & $<$ 0.0001 \\ 
  ti(normalized\_t,duration):gendermale & 19.2208 & 21.7801 & 13.2165 & $<$ 0.0001 \\ 
  s(normalized\_utt\_pos) & 4.7356 & 4.9530 & 113.9360 & $<$ 0.0001 \\ 
  ti(normalized\_t,normalized\_utt\_pos) & 20.2426 & 23.0827 & 4.6701 & $<$ 0.0001 \\ 
  s(bg\_prob\_prev) & 2.9185 & 3.5727 & 44.1743 & $<$ 0.0001 \\ 
  ti(normalized\_t,bg\_prob\_prev):genderfemale & 18.3278 & 21.4065 & 4.6347 & $<$ 0.0001 \\ 
  ti(normalized\_t,bg\_prob\_prev):gendermale & 14.0414 & 17.7218 & 3.2104 & $<$ 0.0001 \\ 
  s(bg\_prob\_fol) & 3.2927 & 3.9600 & 1.6113 & 0.2310 \\ 
  ti(normalized\_t,bg\_prob\_fol):genderfemale & 12.2479 & 15.9275 & 4.9428 & $<$ 0.0001 \\ 
  ti(normalized\_t,bg\_prob\_fol):gendermale & 18.4651 & 21.4256 & 4.3284 & $<$ 0.0001 \\ 
  s(normalized\_t,tonal\_context) & 255.1374 & 324.0000 & 5.8523 & $<$ 0.0001 \\ 
  s(normalized\_t,speaker) & 421.3518 & 494.0000 & 21.9517 & $<$ 0.0001 \\ 
  s(normalized\_t,word) & 304.4643 & 359.0000 & 13.4563 & $<$ 0.0001 \\ 
   \hline
\end{tabular}
\end{adjustbox}
\label{tab:gam}
\end{table}

\begin{figure}[htbp]
    \centering
    \begin{subfigure}[b]{\linewidth}
        \centering
        \includegraphics[width=\linewidth]{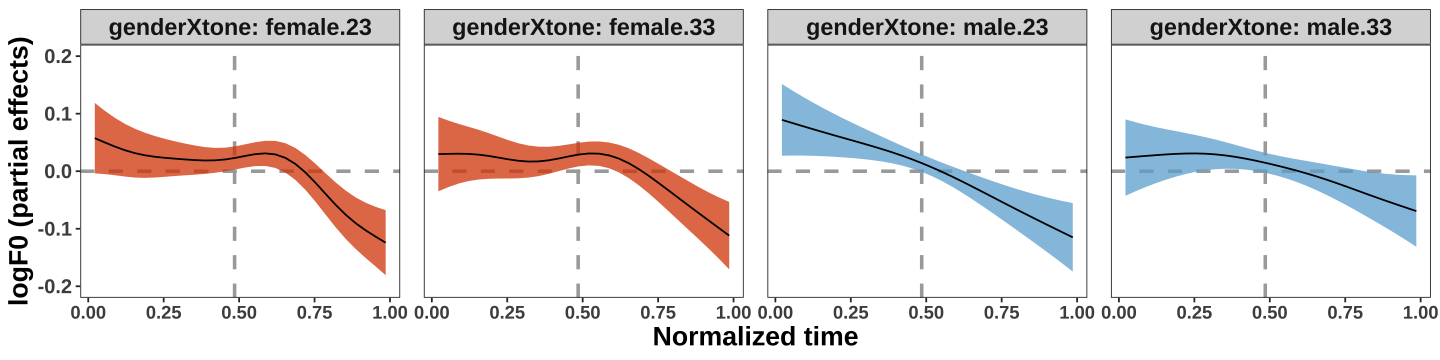}
    \end{subfigure}
    \hfill
    \begin{subfigure}[b]{\linewidth}
        \centering
        \includegraphics[width=\linewidth]{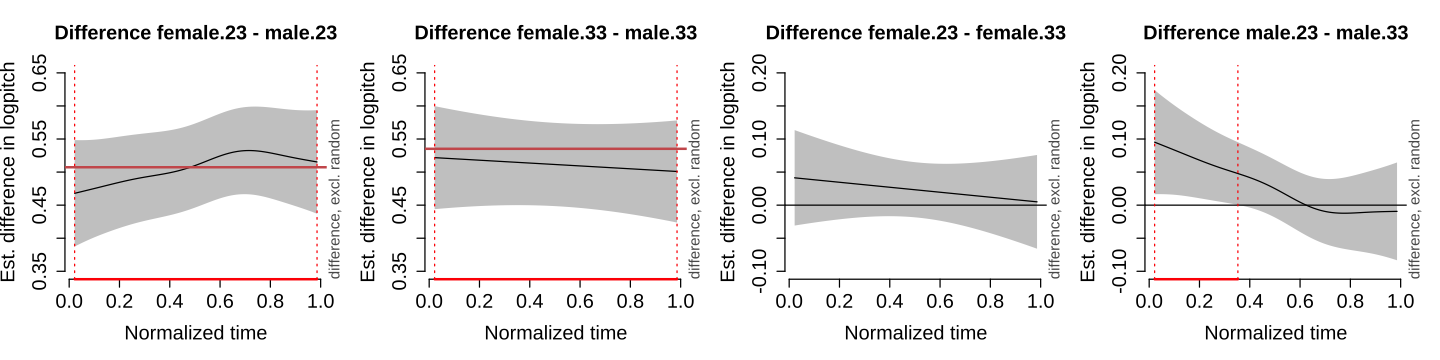}
    \end{subfigure}
    \caption{The partial effects of the three-way interaction of time by gender by tone pattern estimated by the extended GAMM with by-gender modulations.
    }
    \label{fig:genderXtone2}
\end{figure}

The model with by-gender modulations achieves a further model fit improvement compared to the best-fit model reported in Section~\ref{sec:results} by 474.01 AIC units. Nonetheless, these interactions do not change the main conclusions about the complete neutralization of Tone 3 sandhi. Figure~\ref{fig:genderXtone2} visualizes the partial effects of the three-way interaction between time, gender, and tone pattern. Consistent with Figure~\ref{fig:genderXtone}, male and female speakers show a difference in pitch height, but not in the shape of pitch contours. For females, the difference curve between two tone patterns is not well supported. For males, we observe a difference in the beginning of the syllable, but it quickly diminishes to zero. Thus, the inclusion of gender-based modulations could further improves model fit. However, these effects only serve to the fine-tuning of modelling and do not change the primary conclusion of tonal assimilation. 

